 \RequirePackage{amsthm}
\documentclass[pdflatex,sn-mathphys-num]{sn-jnl}

\usepackage{graphicx}%
\usepackage{multirow}%
\usepackage{amsmath,amssymb,amsfonts}%
\usepackage{amsthm}%
\usepackage{mathrsfs}%
\usepackage[title]{appendix}%
\usepackage{xcolor}%
\usepackage{textcomp}%
\usepackage{manyfoot}%
\usepackage{booktabs}%
\usepackage{algorithm}%
\usepackage{algorithmicx}%
\usepackage{algpseudocode}%
\usepackage{listings}%
\usepackage{subcaption}
\usepackage{soul}
\usepackage{fix-cm}

\theoremstyle{thmstyleone}%
\theoremstyle{thmstyletwo}%

\theoremstyle{thmstylethree}%

\begin{document}

\title[Improving Optimization Results with Bandit Networks]{Improving Portfolio Optimization Results with Bandit Networks}


\author{\fnm{Gustavo} \sur{de Freitas Fonseca}}\email{gustavo.fonseca@ga.ita.br}

\author{\fnm{Lucas} \sur{Coelho e Silva}}\email{lucas.coelho@ga.ita.br}

\author{\fnm{Paulo André} \sur{Lima de Castro}}\email{pauloac@ita.br}

\affil{
\orgdiv{Autonomous Computational Systems Lab - LABSCA}\\
\orgname{Aeronautics Institute of Technology (ITA)} \\ 
\orgaddress{\city{São José dos Campos}, \state{São Paulo}, \country{Brazil}}}








\abstract{In Reinforcement Learning (RL), multi-armed Bandit (MAB) problems have found applications across diverse domains such as recommender systems, healthcare, and finance. Traditional MAB algorithms typically assume stationary reward distributions, which limits their effectiveness in real-world scenarios characterized by non-stationary dynamics. This paper addresses this limitation by introducing and evaluating novel Bandit algorithms designed for non-stationary environments. First, we present the \textit{Adaptive Discounted Thompson Sampling} (ADTS) algorithm, which enhances adaptability through relaxed discounting and sliding window mechanisms to better respond to changes in reward distributions. We then extend this approach to the Portfolio Optimization problem by introducing the \textit{Combinatorial Adaptive Discounted Thompson Sampling} (CADTS) algorithm, which addresses computational challenges within Combinatorial Bandits and improves dynamic asset allocation. Additionally, we propose a novel architecture called Bandit Networks, which integrates the outputs of ADTS and CADTS, thereby mitigating computational limitations in stock selection. Through extensive experiments using real financial market data, we demonstrate the potential of these algorithms and architectures in adapting to dynamic environments and optimizing decision-making processes. For instance, the proposed bandit network instances present superior performance when compared to classic portfolio optimization approaches, such as capital asset pricing model, equal weights, risk parity, and Markovitz, with the best network presenting an out-of-sample Sharpe Ratio 20\% higher than the best performing classical model.}




\maketitle


\section{Introduction}\label{sec1}
In the field of Reinforcement Learning (RL), there has been a growing research interest in Multi-Armed Bandit (MAB) problems, a particular problem of RL interpreted as a tabular solution method, where storing transitions does not matter (\citeauthor{Charpentier:2023aa}, \citeyear{Charpentier:2023aa}). Despite their simplicity, these problems have gained attention for their effectiveness in addressing real-world challenges, finding applications ranging from recommender systems (\citeauthor{silva_multi-armed_2022}, 
 \citeyear{silva_multi-armed_2022}), human search behavior (\citeauthor{Nakazato:2024aa}, \citeyear{Nakazato:2024aa}) and information retrieval (\citeauthor{losada_multi-armed_2017}, \citeyear{losada_multi-armed_2017}) to domains like healthcare (\citeauthor{zhou_spoiled_2023}, \citeyear{zhou_spoiled_2023}) and finance (\citeauthor{djallel2020}, \citeyear{djallel2020}). 
 
 Most of the classical MAB framework assumes stationary reward distributions, where the underlying probabilities remain constant over time. However, real-world applications often feature inherently non-stationary environments that may undergo shifts in their probability distributions. In this sense, the need to address non-stationarity arises. One real environment with such behavior is the finance field (\citeauthor{castro_autonomous_2016}, \citeyear{castro_autonomous_2016}), 
where changes in market dynamics demand rapid model responses as to avoid unnecessary risk (\citeauthor{sbrana_nbeats_2023}, \citeyear{sbrana_nbeats_2023} and \citeauthor{Castro2014ModelingAP}, \citeyear{Castro2014ModelingAP}). Similarly, in online advertising, user preferences and behavior may change over time, necessitating adaptive strategies to optimize ad placement and maximize click-through rates. If one intends to use MAB to solve these problems, the non-stationary variants might be a great fit.

In such dynamic contexts, traditional algorithms falter, leading to the need for the development of novel strategies capable of adapting to changing reward structures in real-time. While current literature discusses algorithms that deal with non-stationarity in Bandits problems, existing solutions often encounter limitations regarding the temporal dynamics of policy adaptation, as current formulations exhibit varying degrees of responsiveness to environmental shifts. In this sense, there are opportunities for exploring novel modeling approaches and algorithms, especially toward improved dynamic adaptability of Bandit algorithms under non-stationary conditions.

Particular to finance, the Portfolio Optimization problem emerges as a pertinent application area for MAB solutions. Portfolio Optimization involves selecting and allocating assets to achieve a desirable balance of risk and return. Traditional approaches often rely on static allocation strategies, which may fail to account for changing market conditions effectively. By improving MAB techniques, Portfolio Optimization can benefit from adaptive allocation strategies that dynamically adjust asset weights in response to evolving market dynamics (\citeauthor{Chen:2024aa}, \citeyear{Chen:2024aa}). However, existing literature addressing Portfolio Optimization with MABs remains sparse, highlighting a significant research gap in this domain.

The contributions of this paper are manifold. First, it addresses the challenge of non-stationarity in MAB problems by proposing the Adaptive Discounted Thompson Sampling (ADTS) algorithm. The ADTS algorithm enhances adaptability through relaxed discounting and sliding window mechanisms, allowing it to respond to changes in reward distributions. The algorithm is evaluated through stock picking and portfolio optimization experiments, using historical data from the S\&P 500 index.

Then, building on the ADTS algorithm, this work introduces the Combinatorial Adaptive Discounted Thompson Sampling (CADTS) algorithm for portfolio optimization within the framework of Combinatorial Bandits. The CADTS algorithm addresses the computational challenges associated with combinatorial bandits and aims to optimize decision-making processes in dynamic environments.

Finally, to further enhance the applicability of these algorithms, a novel architecture called Non-Stationary Bandit Networks is proposed. This architecture integrates the outputs of ADTS and CADTS, mitigating biases and improving the robustness of the stock selection process. The effectiveness of these algorithms and architectures is demonstrated through empirical evidence, showcasing their potential in optimizing financial decision-making in non-stationary environments.

In the following sections, we provide a detailed description of our research on non-stationary bandits and the network concepts. The Literature Review in Section 2 covers recent work about non-stationary bandits and the practical usage of MABs in Portfolio Optimization. In Section 3, we present a detailed description of the novel non-stationary bandit algorithm, Adaptive Discounted Thompson Sampling (ADTS), its combinatorial bandit variant, Combinatorial Adaptive Discounted Thompson Sampling (CADTS) and how they together constitute original architectures called Bandit Networks aimed to solve the Portfolio Optimization problem using historical daily price of the Standard and Poor's (S\&P) stocks. Section 4 presents the experimental setup, aiming to evaluate the ADTS and the bandit network instances applied to financial market data of a set of Standard and Poor's stocks. Section 5 presents our findings through the experiments, including the stock picking, portfolio optimization, and portfolio optimization robustness experiments. In Section 6, we present a comprehensive discussion of the research and introduce practical implications of the ADTS and the bandit networks on finance. Finally, we conclude our study by summarizing the contributions and implications for future research.

\section{Literature Review}\label{sec2}
\label{sec:related_work_main}

The goal of this research is to improve the practical usage of Multi-Armed Bandits (MAB) in changing environments such as finance. This section explores the forefront advancements of Non-Stationary Bandit algorithms and outlines some practical applications of bandit algorithms in finance.

\subsection{Non-Stationary Bandits}
\label{sec:non_stat_bandits}
\label{section:nonstat_bandits}

Unlike traditional stationary bandit settings, where rewards associated with each action remain constant throughout the learning process, non-stationary bandit problems arise in scenarios where the underlying environment is subject to stochastic and agent-independent changes over time, leading to variations in the reward distribution. (\citeauthor{allesiardo:hal-01575000}, \citeyear{allesiardo:hal-01575000}).

The need to address non-stationarity arises in real-world applications where the environment is inherently dynamic and may undergo unpredictable shifts. In finance, stock prices and returns are constantly changing, demanding fast changes to avoid unnecessary risks. Similarly, in online advertising, user preferences and behavior may change over time, necessitating adaptive strategies to optimize ad placement and maximize click-through rates. \citeauthor{lattimore2020bandit} (\citeyear{lattimore2020bandit}) argues that the process of building a non-stationary bandit variant typically occurs by applying discounts or sliding windows to pre-existing stationary policies. These artifices dynamically augment the exploration components and prevent the algorithm from being locked into a local minimum.

\citeauthor{raj2017taming} (\citeyear{raj2017taming}) have introduced the Discounted Thompson Sampling (D TS) philosophy, to continuously increase the variance of the prior distribution and maintain exploration over time, mitigating the effect of past observations. The work also introduced the optimistic version of D TS, the so-called Discounted Optimistic Thompson Sampling (DOTS). In DOTS, the samples are forced to have at least its expected value, thus increasing the arms’ exploitative value. The dTS and dOTS were challenged only in synthetic data, although showcased a good margin of regret in slow and fast varying environments, compared to other algorithms, such as the Classical Thompson Sampling \cite{thompson1933likelihood}.


\citeauthor{Trov2020SlidingWindowTS} (\citeyear{Trov2020SlidingWindowTS}) presented the Sliding-Window Thompson Sampling for non-stationary MAB settings. As the name suggests, it adapts the sampler to a hot trace by inspecting past successes and failures given a sliding window hyper-parameter. This work provides regret upper bounds for dynamic pseudo-regret in different scenarios. Empirical evidence showed that sw-TS outperforms existing algorithms in non-stationary settings.

Following this path, \citeauthor{cavenaghi2021nonstationary} (\citeyear{cavenaghi2021nonstationary}) introduces a new Thompson Sampling variant called f-Discounted-Sliding-Window Thompson Sampling (f-dsw TS) to address concept drift problems. In this case, the work combines both the concepts of discounts and sliding windows. The discount factor adjusts the choices of a historical sampler while the sliding window walks through a short-term sampler, that is processed in parallel. These two samplers are instantiated by each arm using an aggregation function $f(.)$. The aggregation function can compare both historical and short-term samplers based on three types of approaches: i) pessimistic, the minimum between each sampler $(f=min)$, ii) optimistic, the maximum between each sampler $(f=max)$ and iii) the average of the two samplers  $(f=mean)$. The work conducts experiments in synthetic and real-world environments and compares f-dsw TS with stationary and non-stationary TS baselines. Based on the simulations, the f-dsw TS algorithm outperforms baselines in synthetic environments. The pessimistic version $(f=min)$ is most effective in real-world data.

From the frequentist perspective, \citeauthor{garivier2011nonstatucb} (\citeyear{garivier2011nonstatucb}) explores the Discounted Upper Confidence Bound (D UCB) and Sliding Window Upper Confidence Bound (SW-UCB) variants of UCB. The work establishes upper and lower bounds for regret in changing environments and points out that these policies adapt well to non-stationary environments. \citeauthor{pmlr-v89-cao19a} (\citeyear{pmlr-v89-cao19a}) presents an innovative M-UCB algorithm with near-optimal regret bounds, integrating change detection with traditional UCB methods. Experimental comparisons confirm its superior performance. \citeauthor{Liu2017ACB} (\citeyear{Liu2017ACB}) introduces CD-UCB policies, including CUSUM-UCB and PHT-UCB, showcasing regret reduction across synthetic and real datasets.

While the covered state-of-the-art Bandit policies provide attempts at tackling the concept drift in non-stationary settings, there remain unsolved problems regarding the proposition of novel non-stationary Bandit policies, especially toward improved dynamic adaptability of Bandit algorithms in such conditions.

\subsection{Applications of Bandits in Portfolio Optimization}
\label{sec:bandits_finance}
To fulfill our objective of providing a comprehensive overview of the applications of bandits in finance, we explore some applications in the field of finance, ranging from portfolio optimization to high-frequency trading strategies.

Portfolio Optimization is a crucial problem in finance, aiming to allocate assets to achieve optimal returns while managing risks effectively. The application of Reinforcement Learning (RL) techniques, including Bandit algorithms, has garnered significant attention in recent years (\citeauthor{wang2019large}, \citeyear{wang2019large}). Some researchers argue that simpler online learning algorithms like bandits can effectively address the allocation problem (\citeauthor{li2014online}, \citeyear{li2014online}).

Stochastic multi-armed Bandit models, which address the exploration-exploitation trade-off, offer a natural framework for sequential decision-making under uncertainty, making them suitable for portfolio selection. By incorporating risk awareness and employing optimal policies, \citeauthor{huo2017risk} (\citeyear{huo2017risk}) have aimed to strike a balance between risk and return in portfolio construction. Similarly, bandit algorithms have been utilized to exploit correlations among assets, leading to the development of effective online portfolio selection strategies (\citeauthor{shen2015portfolio}, \citeyear{shen2015portfolio}).

Classic portfolio optimization models, such as Markowitz's mean-variance optimization (\citeauthor{markovitz_1952}, \citeyear{markovitz_1952}), face challenges in parameter estimation and applicability across different market conditions. In contrast, bandit-based strategies offer flexibility and adaptability, particularly in environments where traditional models may falter. By treating different portfolio strategies as strategic arms in a multi-armed bandit setup, \citeauthor{zhu2019adaptive} (\citeyear{zhu2019adaptive}) has sought to maximize rewards through a judicious balance of exploration and exploitation.

In summary, the literature on real applications of bandits in portfolio optimization problems is still incipient, which highlights opportunities to contribute to the research by unifying these two different domains.

\section{Proposal}\label{sec3}

In this Section, we present a detailed description of the novel non-stationary bandit algorithm, Adaptive Discounted Thompson Sampling (ADTS), its combinatorial bandit variant, Combinatorial Adaptive Discounted Thompson Sampling (CADTS) and how they together constitute original architectures called Bandit Networks aimed to solve the Portfolio Optimization problem using historical daily price of the Standard and Poor's (S\&P) stocks. We empirically evaluate the performance of the proposed algorithms and network architectures based on the experiments presented in the subsequent sections.

\subsection{Adaptive Discounted Thompson Sampling}
\label{sec:rf_dsw_ts}
We introduce the Adaptive Discounted Thompson Sampling (ADTS), a Thompson Sampling (TS) (\citeauthor{thompson1933likelihood}, \citeyear{thompson1933likelihood}) variant aimed to deal with non-stationary environments more efficiently. 

The TS algorithm tracks the rewards history $X_t^k$ for each arm $k$ using a Bernoulli distribution, denoted as $\mathcal{B}(\alpha_k, \beta_k)$, which has the parameters $\alpha$ and $\beta$. Physically, $\alpha$ can be interpreted as the cumulative success counts while $\beta$ works as the cumulative failure counts. In that sense, the distribution $\mathcal{B}(\alpha_k, \beta_k)$ yields the expected success value by pulling each arm $k$. The classical TS updating is governed by the expression: 

\begin{equation}
    \mathcal{B}(\alpha_k, \beta_k) = 
    \begin{cases}
        \mathcal{B}(\alpha_k, \beta_k), & \text{if } I_t \neq k \\
        \mathcal{B}(\alpha_k + X_t^k, \beta_k + 1 - X_t^k), & \text{if } I_t = k 
    \end{cases}
\end{equation}

where $I_t$ is the selected arm at step $t$.

In the language of bandits, the regret $R(t)$ represents the cumulative learning error. It quantifies the difference between an always optimal choice, or oracle (\citeauthor{NIPS2014_903ce922}, \citeyear{NIPS2014_903ce922}), and the sub-optimal choices by some bandit policy:

\begin{equation}
\label{eq:regret}
    R(t) = \sum_{t=1}^{T}X_t^* - \mathbb{E}{\left[\sum_{t=1}^{T}X_t^k\right]}
\end{equation}

where $X_t^*$ is the optimal reward at step $t$.

The regret $R(t)$ measure can be used to compare and contrast different bandit policies. In non-stationary environments such as the financial stock markets, where the stock returns distributions are changing, classical bandit algorithms tend to show higher regret values, as they usually get stuck into some local optimum arm. To minimize this problem, the non-stationary bandit variants appeared.

The ADTS algorithm (\citeauthor{rfdswts_2024}, \citeyear{rfdswts_2024}), adapted from (\citeauthor{cavenaghi2021nonstationary}, \citeyear{cavenaghi2021nonstationary}), relaxes the application of the discount factor by applying it only for the selected arm $I_t$, instead of applying it for all of the arms. On the other hand, we keep intact the construction of what we interpret as the short-term memory of the policy, by applying the sliding window approach, and then comparing both discounted and short-term samples with the aggregation function $f(.)$ for each arm. 

\begin{figure}[ht]
\centering
\includegraphics[width=0.95\textwidth]{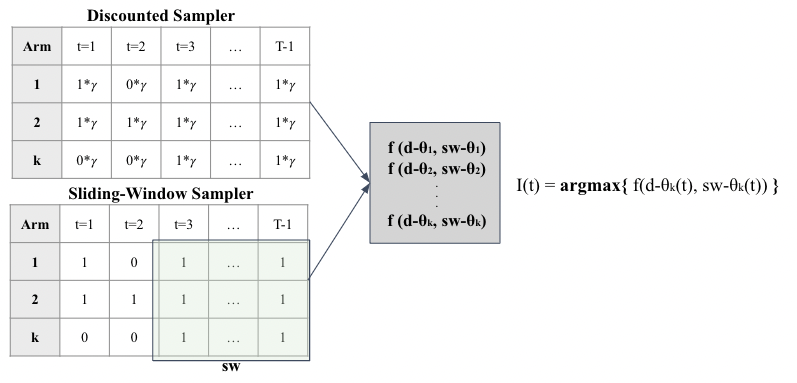}
\caption{ADTS Selection Diagram.}
\label{fig:4.1_rfswdts_diagram}
\end{figure}

More formally, the discount factor $\gamma \in \left(0, 1\right]$ gradually diminishes the impact of past observations in the historic trace. The short-term trace, represented as $\breve{\mathcal{B}}(\alpha_k^n, \beta_k^n)$, tracks the recent rewards by applying a sliding window $w$. The mix between the historical and hot traces components is performed before the arm is played at step $t$. For each arm $k$, the algorithm computes an aggregated score $S_k(t)$, as:

\begin{equation}
    S_k(t) = f(\theta_k(t),\breve{\theta}_k(t))  
\end{equation}

where $f(.)$ is the aggregation function defined for the algorithm ($min$, $avg$, $max$), $\theta_k(t)$ is a sample from the historic trace distribution $\mathcal{B}(\alpha_k, \beta_k)$, $\breve{\theta}_k(t)$ is a sample from the short-term trace distribution $\breve{\mathcal{B}}(\alpha_k^w, \beta_k^w)$ at step $t$ for arm $k$. Finally, ADTS chooses which arm to play at step $t$ as $I_t=argmax(S_k(t))$. Figure~\ref{fig:4.1_rfswdts_diagram} depicts the arms-pulling process when computing in parallel the short-term and historical traces.

Algorithm~\ref{alg:alg1} outlines the ADTS strategy. In lines (2–5), for each arm, we sample a reward estimate from both historic and short-term distributions. In line (6) we apply the aggregation function $f (.)$ to select one of the two estimates (or a mix of them) and choose the arm with the highest aggregated score. We pull the arm and observe the reward at time t (i.e., rt) in line (7). In line 9, we apply the discount factor only for the selected arm $k = I_t$.

\begin{algorithm}[H]
\caption{\textit{Adaptive Discounted Thompson Sampling TS}.}
\label{alg:alg1}
\textbf{Input:} \\
$k=|\mathcal{K}|\geq2$ number of arms \\ 
$\gamma \in (0, 1]$ Discount factor\\ 
$w \in [1, T]$ Sliding-window size
\begin{algorithmic}[1]
\For{$t = 1,2,\ldots,T$}
    \For{$k = 1,2,\ldots,K$}
        \State $\theta_k(t) = \mathcal{B}(\alpha_k + 1, \beta_k + 1)$
        \State $\breve{\theta}_k(t) = \breve{\mathcal{B}}(\alpha_k^w, \beta_k^w)$
    \EndFor
    \State Play arm $I(t) = \arg\max_k(f(\theta_k(t),\breve{\theta}_k(t)))$
    \State Observe reward $r_t$
    \State $X_t = 1 \quad \text{if} \quad r_t = r_t^* \quad \text{else} \quad 0$
    \State Update $\mathcal{B}(\alpha_k, \beta_k)$ as:
    \State $
        \mathcal{B}(\alpha_k, \beta_k) = 
        \begin{cases}
            (\alpha_k, \beta_k), & \forall k \neq I_t \\            
            \gamma(\alpha_{I_t}, \beta_{I_t}) + (X_t, 1-X_t)
        \end{cases}
    $
    \State Update $\breve{\mathcal{B}}(\alpha_k^w, \beta_k^w)$, where $k = I_t$, with the last $w$ rewards taken for arm $k$
\EndFor
\end{algorithmic}
\end{algorithm}

\subsection{Combinatorial Adaptive Discounted Thompson Sampling}
\label{sec:crf_dsw_ts}
In this section, we extend the ADTS algorithm to a combinatorial bandit problem, originating the Combinatorial Adaptive Discounted Thompson Sampling Thompson Sampling (CADTS).

For conducting the Portfolio Optimization problem in the context of Bandits, we combined ADTS with the Combinatorial Bandits formulation proposed by \citeauthor{chen_2013_combinatorial} (\citeyear{chen_2013_combinatorial}). Theoretically, each stock can have infinite possible weight values $w_k$, which can lead our CADTS to dimensionality issues. To avoid that, we construct our feasible portfolio weights combinations \textbf{}(superarms) by building an array of discrete weights $pw_{k}$ for each stock $k$, given the total number of stocks $K$.

\begin{equation}
    \label{eq:possible_weights}
    pw[k, :] = \begin{bmatrix}
        0 & 1s & 2s & 3s & ... & 1
    \end{bmatrix}
\end{equation}

where $s = \frac{1}{2K}$ is the minimum weight step value.

Then, we construct the possible weights matrix $PW$, where the rows represent each stock and the columns are the possible weights for each stock defined in equation~\ref{eq:possible_weights}. To create the super-arms, we run all the possible weight combinations between the stocks and possible weights in the $PW$ matrix $s.t. \sum_{k=1}^{K} w_k(t) = 1$:

\begin{equation}
    \label{eq:possible_weights_matrix}
    PW = \begin{bmatrix}
        \multirow{2}{*}{$\text{Stock}_1$}\\
        & 0 & 1s & 2s &\cdots & 1 \\
        \multirow{2}{*}{$\text{Stock}_2$}\\
        & 0 & 1s & 2s &\cdots & 1 \\
        \vdots & \vdots & \vdots & \ddots & \vdots \\
        \multirow{2}{*}{$\text{Stock}_k$}\\
        & 0 & 1s & 2s &\cdots & 1 \\
        \end{bmatrix}
\end{equation}

Algorithm~\ref{alg:alg2} outlines the complete pseudo-code for the Combinatorial Adaptive Discounted Thompson Sampling TS (CADTS). Lines (1-5) describe the procedure to generate the feasible superarms $\mathcal{S}$ containing the combinations of weights for each stock. From lines (7-19) we repeat the non-stationary bandit problem outlined in Algorithm~\ref{alg:alg1}.

\begin{algorithm}[H]
\caption{\textit{Combinatorial Adaptive Discounted Thompson Sampling TS}.}
\label{alg:alg2}
\textbf{Input:} \\
$K$ Number of arms\\
$s = \frac{1}{2K}$ Minimum weight step value\\ 
$\gamma \in (0, 1]$ Discount factor\\ 
$w \in [1, T]$ Sliding-window size \\
$n$ Number of arms inside each superarm $S$ \\
\begin{algorithmic}[1]
\State{\textbf{Generate the portfolio feasible super arms ($\mathcal{S}$ given $K$ and $s$}}
\For{$k = 1,2,\ldots,K$}
    \State $pw[k, :] = \begin{bmatrix}
        0 & ks & 2ks & 3ks & ... & 1
    \end{bmatrix}$
\EndFor
\State $\mathcal{S} = \begin{bmatrix}
    Combinations(w_{k, i}) & | & \sum_{k=1}^{K} pw_{k, i} = 1
\end{bmatrix}$
\\
\State{\textbf{Run ADTS}}
\For{$t = 1,2,\ldots,T$}
    \For{$S = 1,2,\ldots,\mathcal{S}$}
        \State $\theta_S(t) = \mathcal{B}(\alpha_S + 1, \beta_S + 1)$
        \State $\breve{\theta}_S(t) = \breve{\mathcal{B}}(\alpha_S^w, \beta_S^w)$
    \EndFor
    \State Play arm $I_t = \arg\max_S(f(\theta_S(t),\breve{\theta}_S(t)))$
    \State Observe the portfolio reward $r_t = \sum_{k=1}^{K} w_k(t) r_k(t)$ 
    \State $X_t = 1 \quad \text{if} \quad r_t = r_t^* \quad \text{else} \quad 0$
    \State Update $\mathcal{B}(\alpha_S, \beta_S)$ as:
    \State $
        \mathcal{B}(\alpha_S, \beta_S) = 
        \begin{cases}
            (\alpha_S, \beta_S), & \forall k \neq I(t) \\            
            \gamma(\alpha_{I_t}, \beta_{I_t}) + (X_t, 1-X_t)
        \end{cases}
    $
    \State Update $\breve{\mathcal{B}}(\alpha_S^w, \beta_S^w)$, where $S = I_t$, with the last $w$ rewards taken for super arm $S$
\EndFor
\end{algorithmic}
\end{algorithm}

\subsection{Bandit Networks}
\label{sec:non_bandits_networks}
We demonstrated in (\citeauthor{rfdswts_2024}, \citeyear{rfdswts_2024}) that the ADTS algorithm efficiently selects the best arm in a changing environment. Heavily inspired by the Neural Networks philosophy, we introduce a novel approach called Bandit Networks. It connects between layers of non-stationary bandits policies such as ADTS and CADTS. In this section we propose two different architectures to solve the Portfolio Optimization problem: i) Non-Stationary Bandit with CADTS Network and ii) Two-layer ADTS Network.

\subsubsection{Non-Stationary Bandit with CADTS Network}
\label{sec:comb_network}

\begin{figure}[ht]
\centering
\includegraphics[width=0.95\textwidth]{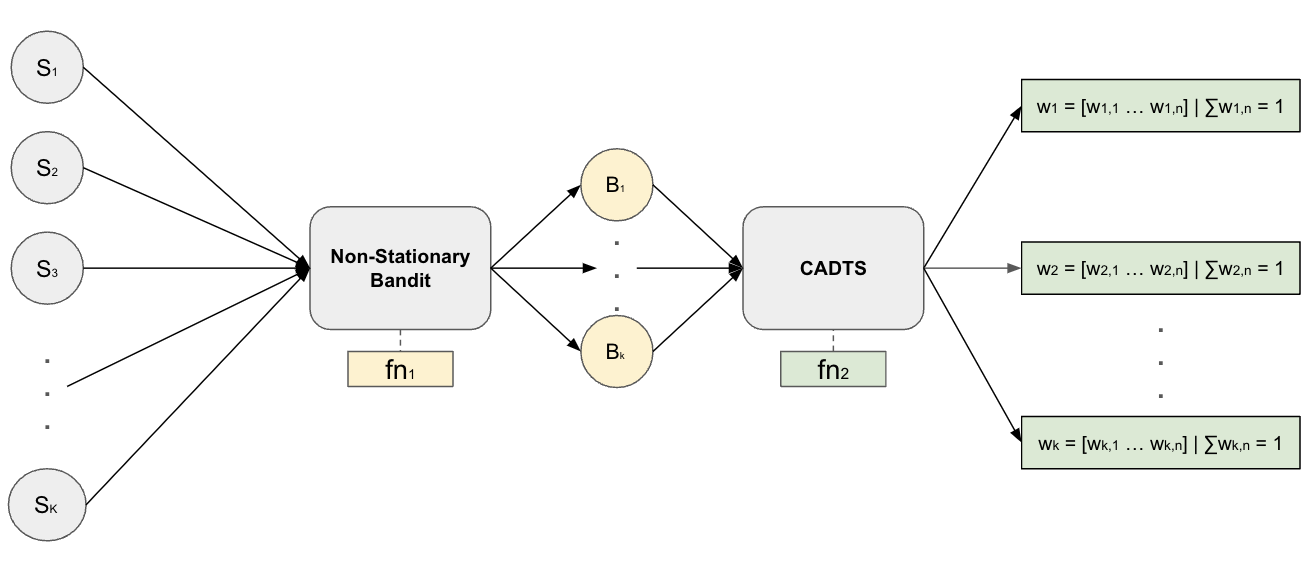}
\caption{Non-Stationary Bandit with CADTS Network architecture}
\label{fig:4.3_bandits_network}
\end{figure}

Figure~\ref{fig:4.3_bandits_network} displays the Non-Stationary Bandit with CADTS Network architecture. In the first layer, the non-stationary Bandit policy (ADTS, D TS, SW UCB, or any other)  receives $S_1, S_2, ... S_K$, the complete universe of stocks. More than selecting the best stock at time step $t$, the role of the non-stationary Bandit policy is to provide the second layer the rank of the $k < K$ best stocks, based on the reward function $fn_1$, colored in yellow in the diagram. The function $fn$ can be constructed to select stocks based on historical or sliding-window cumulative returns, momentum, or risk-adjusted returns, such as the Sharpe Index.

Having the $k$ best stocks, the CADTS generates the portfolio feasible weights combinations $s.t. \sum_{k=1}^{K} pw_{k, i} = 1$, as described in Algorithm~\ref{alg:alg2}. The policy is accountable for selecting at time step $t$ the best weight combination (super arm) that maximizes its reward function $fn_2$, colored in green. Similarly, the second layer objective function can also be constructed to select stocks based on a financial metric, whether the same as $fn_1$ or a different one.

\subsubsection{Two-layer ADTS Network}
\label{sec:2_layer_network}

\begin{figure}[ht]
\centering
\includegraphics[width=0.95\textwidth]{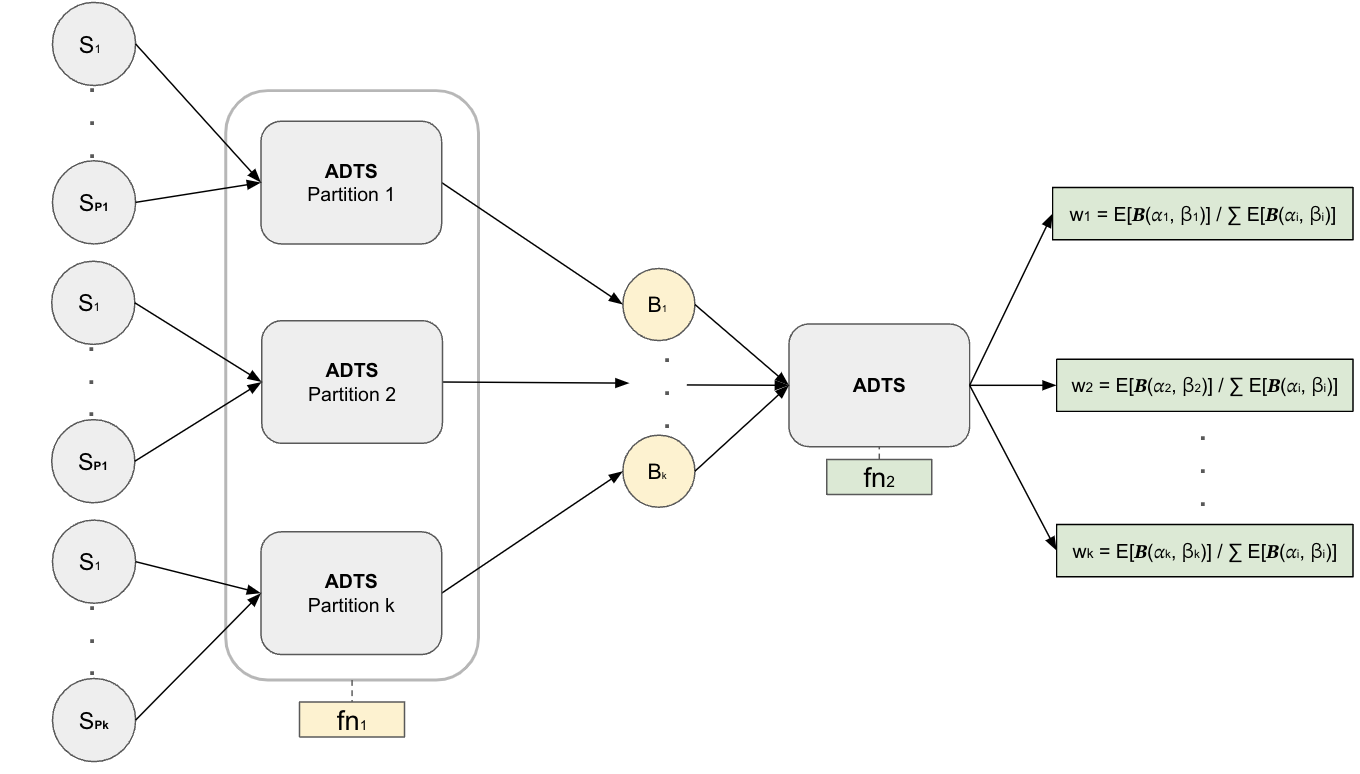}
\caption{Two-layer ADTS Network architecture}
\label{fig:4.4_two_tage_rf}
\end{figure}

We present an alternate to the Non-Stationary Bandit with CADTS Network architecture. Figure~\ref{fig:4.4_two_tage_rf} displays the Two-layer ADTS Network. In the first layer, we partition the total stocks universe into $k$ parts. For each partition, we run an ADTS to filter the stocks universe using the reward function $fn_1$, colored in yellow in the diagram.

Given the $k$ best stocks to the second layer, we bolt another ADTS to learn the $k$ best stocks hierarchy given another reward function $fn_2$, colored in green. The portfolio weights are generated by normalizing the expected success values of each $k$ Bernoulli distribution:

\begin{equation}
    w_i = \frac
    {\mathbb{E}[\mathcal{B}(\alpha_i, \beta_i)]}
    {\sum_{i=1}^{k} {\mathbb{E}[\mathcal{B}(\alpha_i, \beta_i)]}}
\end{equation}

\section{Experimental Setup}\label{sec4}
The experiments designed in this paper aim to demonstrate the practical usefulness of the proposed ADTS and CADTS algorithms and their connection to form the Bandit Networks in a real environment provided by a set of daily returns of S\&P stocks.

To achieve this goal, we divide the experiments into three phases. In the first phase, we investigate if the ADTS effectively selects the best S\&P stock in the so-called Stock Picking problem. Next, we evaluate the performance of different Bandit Networks instances in the Portfolio Optimization Problem given a set of S\&P stocks. Finally, we check the Bandit Networks instance's results robustness by removing a set of high-performing stocks. 

In this section, we present the selected set of S\&P stocks and the collected daily market data, describe each of the three experiments, and present the benchmarks used to compare the results.

\subsection{Market Data and Problem Definition}
\label{sec:sp_market_data}
We submit the non-stationary bandit's policies to a real-world problem by selecting a set of 44 stocks within the S\&P index, taking their historical prices from April 2020 to July 2024. These stocks behave as our arms in a bandit problem context. Given the historical series of daily returns $\begin{bmatrix} r_0, r_1, r_2, ..., r_t \end{bmatrix}$, where $t$ represent each time step, the objective is to maximize the future rewards $\begin{bmatrix} X_{t+1}, X_{t+2}, X_{t+3}, ..., X_T \end{bmatrix}$ either for a unique stock or a portfolio of stocks. The bandit reward function is defined by $F(\begin{bmatrix} r_{{w_f}-t}, ..., r_{{w_f}-2}, r_{w_f}, r_{w_f} \end{bmatrix})$, where $F$ is a financial metric such as cumulative Returns, Sharpe Index, Sortino Ratio, etc and $w_f$ represents the window length in case of applying sliding window to the financial function. If the historical financial function is desired at each time step $t$, the sliding window becomes infinite and no hyperparameter $w_f$ is necessary.

\begin{figure}[H]
\centering
\includegraphics[width=0.9\textwidth]{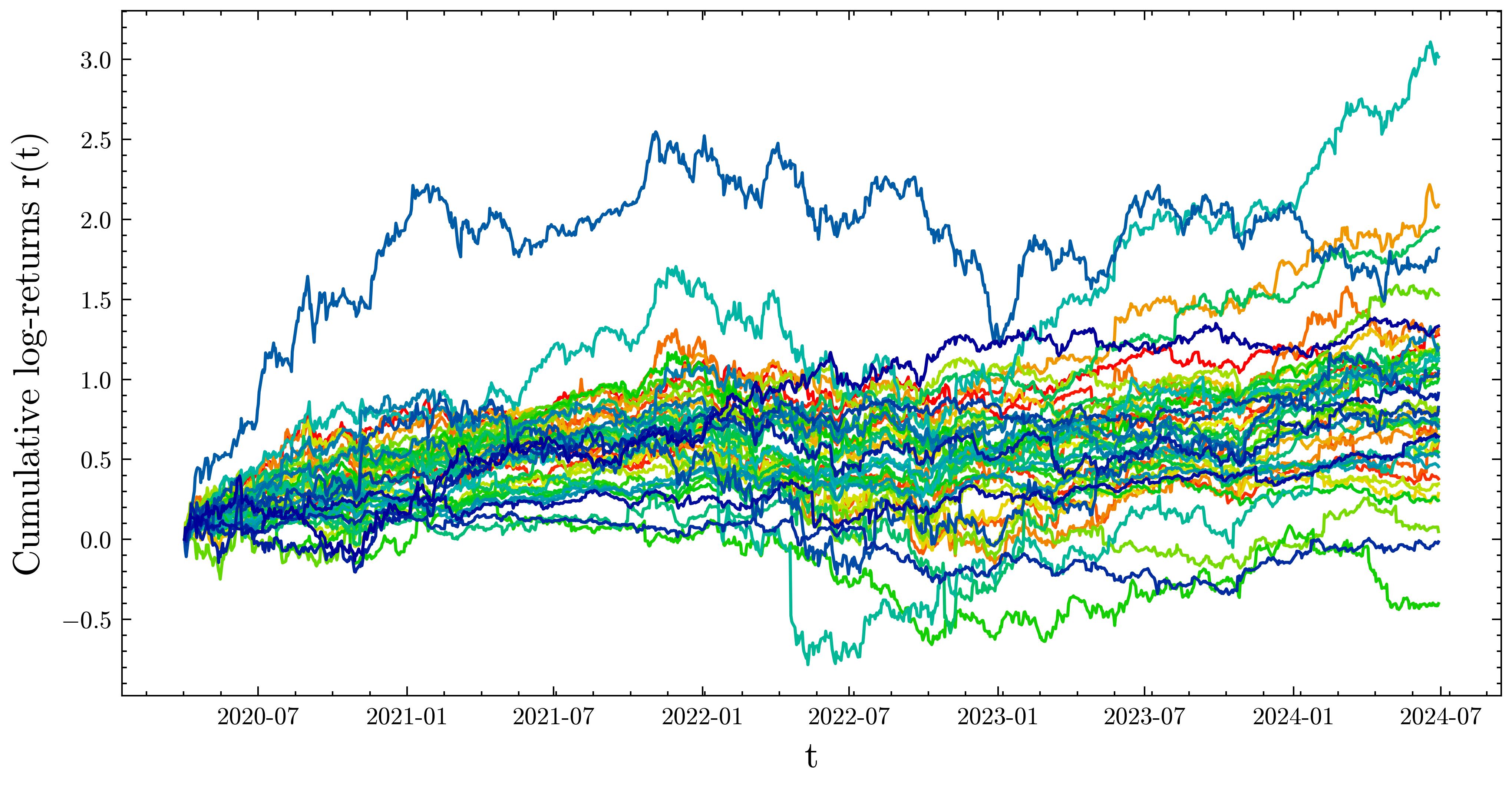}
\caption{Selected S\&P stocks logarithmic cumulative daily returns}
\label{fig:sp_log_returns}
\end{figure}

\begin{figure}[H]
\centering
\includegraphics[width=0.9\textwidth]{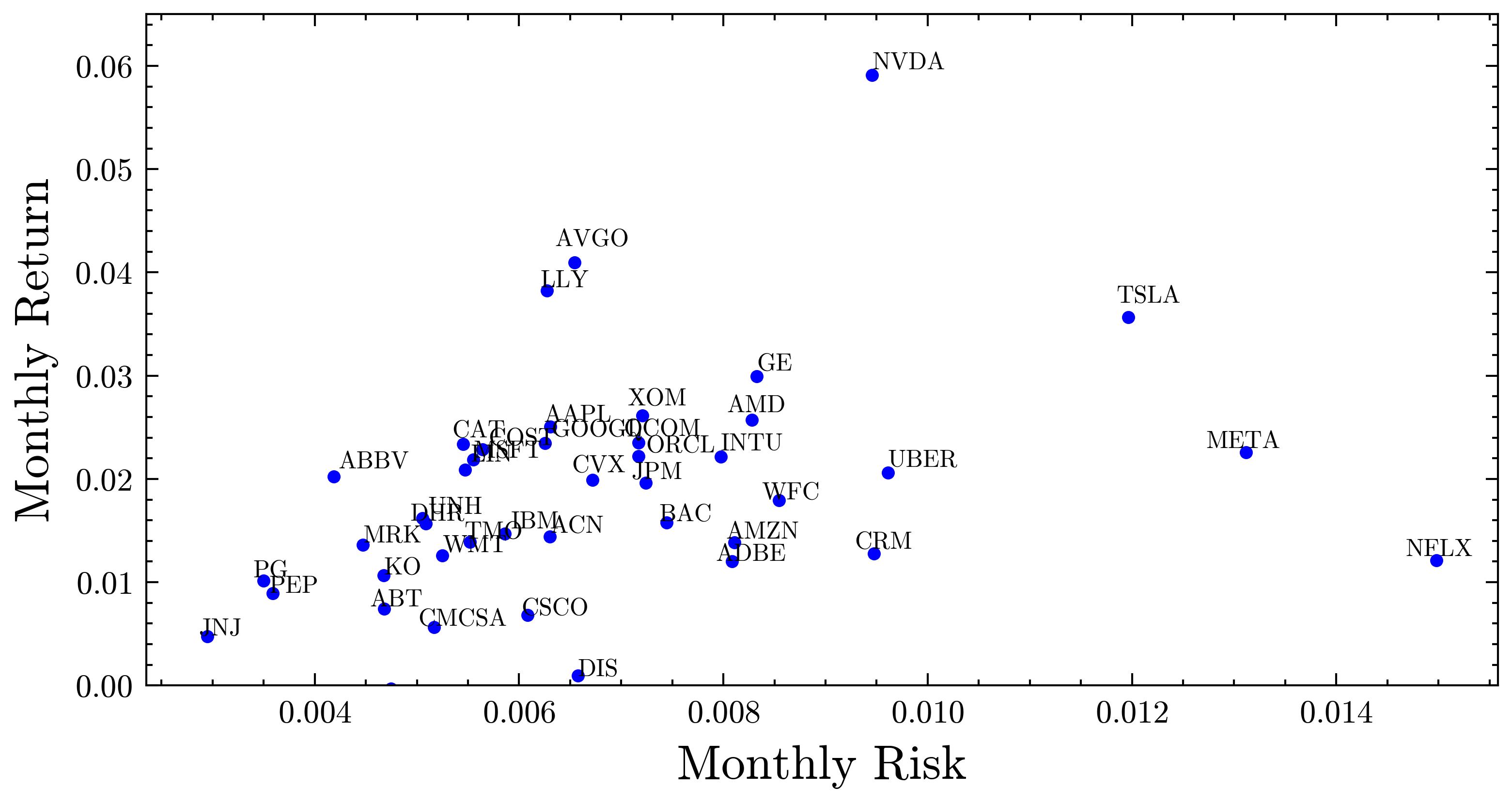}
\caption{Monthly log-return and risk for each selected stock}
\label{fig:sp_risk_return}
\end{figure}

The arms' logarithmic cumulative daily returns are shown in Figure~\ref{fig:sp_log_returns} and the monthly log returns and risks are displayed in Figure~\ref{fig:sp_risk_return}. Table~\ref{tab:sp_risk_return} summarizes the monthly returns and risks.

\begin{table}[ht]
\caption{Stocks Monthly Risks and Returns}
\centering
\begin{tabular}{lclc}
\hline
Stock Symbol & Monthly Return $\pm$ Monthly Risk & Stock Symbol & Monthly Return $\pm$ Monthly Risk \\
\hline
NVDA & $0.059 \pm 0.009$ & UNH & $0.016 \pm 0.005$ \\
AVGO & $0.041 \pm 0.007$ & BAC & $0.016 \pm 0.007$ \\
LLY & $0.038 \pm 0.006$ & DHR & $0.016 \pm 0.005$ \\
TSLA & $0.036 \pm 0.012$ & IBM & $0.015 \pm 0.006$ \\
GE & $0.030 \pm 0.008$ & ACN & $0.014 \pm 0.006$ \\
XOM & $0.026 \pm 0.007$ & TMO & $0.014 \pm 0.006$ \\
AMD & $0.026 \pm 0.008$ & AMZN & $0.014 \pm 0.008$ \\
AAPL & $0.025 \pm 0.006$ & MRK & $0.014 \pm 0.004$ \\
QCOM & $0.024 \pm 0.007$ & CRM & $0.013 \pm 0.009$ \\
GOOGL & $0.023 \pm 0.006$ & WMT & $0.013 \pm 0.005$ \\
CAT & $0.023 \pm 0.005$ & NFLX & $0.012 \pm 0.015$ \\
COST & $0.023 \pm 0.006$ & ADBE & $0.012 \pm 0.008$ \\
META & $0.023 \pm 0.013$ & KO & $0.011 \pm 0.005$ \\
ORCL & $0.022 \pm 0.007$ & PG & $0.010 \pm 0.003$ \\
INTU & $0.022 \pm 0.008$ & PEP & $0.009 \pm 0.004$ \\
MSFT & $0.022 \pm 0.006$ & ABT & $0.007 \pm 0.005$ \\
LIN & $0.021 \pm 0.005$ & CSCO & $0.007 \pm 0.006$ \\
UBER & $0.021 \pm 0.010$ & CMCSA & $0.006 \pm 0.005$ \\
ABBV & $0.020 \pm 0.004$ & JNJ & $0.005 \pm 0.003$ \\
CVX & $0.020 \pm 0.007$ & DIS & $0.001 \pm 0.007$ \\
JPM & $0.020 \pm 0.007$ & VZ & $-0.000 \pm 0.005$ \\
WFC & $0.018 \pm 0.009$ & INTC & $-0.008 \pm 0.007$ \\
\hline
\label{tab:sp_risk_return}
\end{tabular}
\end{table}



\subsection{Stock Picking Experiment}
\label{sec:s_p_stock}
In this experiment, we aim to apply the proposed ADTS algorithm to the stock picking problem, given the set of the S\&P stocks from~\ref{sec:sp_market_data}. To provide benchmark comparisons to our non-stationary bandit variant, we invoke the bandit algorithms listed below as Table~\ref{tab:sp_stock_picking_setup} summarizes the complete experiment setup.

\begin{itemize}
    \item Classical Thompson Sampling: Classical TS (\citeauthor{thompson1933likelihood}, \citeyear{thompson1933likelihood});
    \item f-Discounted-Sliding-Window Thompson Sampling: F-DSW TS (\citeauthor{cavenaghi2021nonstationary}, \citeyear{cavenaghi2021nonstationary});
    \item Discounted Thompson Sampling: D-TS (\citeauthor{raj2017taming}, \citeyear{raj2017taming});
    \item UCB-1 (\citeauthor{auer2002ucb}, \citeyear{auer2002ucb}).
    \item Discounted UCB: D-UCB (\citeauthor{garivier2011nonstatucb}, \citeyear{garivier2011nonstatucb});
    \item Sliding-Window UCB: SW-UCB (\citeauthor{garivier2011nonstatucb}, \citeyear{garivier2011nonstatucb});
\end{itemize}

\begin{table}[h]
\caption{S\&P Stock Picking - Experiment Setup}
\centering
\begin{tabular}{llccl}
\hline
Algortihm & Bandit Family & Reward Function & Window Length ($w_f$) & Hyper-Parameters\\
\hline
Classical TS  & - & Mean Return & 100 days & - \\
ADTS & TS & Mean Return & 100 days & $\gamma=0.9$; $f=mean$; $w=100$ \\
F-DSW TS  & TS & Mean Return & 100 days & $\gamma=0.99$; $f=mean$; $w=100$ \\
D TS  & TS & Mean Return & 100 days & $\gamma=0.99$ \\
 & & & & \\
UCB-1  & - & Mean Return & 100 days & -\\
D UCB & UCB-1 & Mean Return & 100 days & $\gamma=0.1$\\
SW UCB & UCB-1 & Mean Return & 100 days & $w=50$\\
\hline
\label{tab:sp_stock_picking_setup}
\end{tabular}
\end{table}


Next, we investigate the financial metrics of the experimented policies (Total Return, Sharpe Ratio, Drawdown, Win Rate and Sortino Ratio), comparing with each other and with the S\&P Index.

Finally, we finish the experiment investigating the drift effect on each bandit policy by applying an artificial shock to the best performing ticker, the NVDA. We are interested to investigate how does a shock impact on each policy with respect to their financial metrics distributions.

\subsection{Portfolio Optimization Experiment}
\label{sec:s_p_portfolio}
For conducting the Portfolio Optimization for the S\&P stocks presented in Section~\ref{sec:sp_market_data}, we apply different Bandit Networks instances. Table~\ref{tab:sp_portfolio_setup} summarizes the applied Bandit Networks in the experiments, describing their architectures, algorithms, and rewards functions for each layer and the portfolio size $k$.

In the experiments, we analyze the learning behavior of the network instances in terms of cumulative regret and use financial metrics to compare with the following benchmarks:

\begin{itemize}
    \item Capital Asset Pricing Model (CAPM) - \citeauthor{capm_2004} (\citeyear{capm_2004});
    \item Portfolio Theory - \citeauthor{markovitz_1952} (\citeyear{markovitz_1952});
    \item Risk Parity;
    \item Equal Weights;
    \item S\&P Index.
\end{itemize}

\begin{table}[h]
\caption{S\&P Portfolio Optimization - Experiment Setup}
\centering
\begin{tabular}{ccllllc}
\hline
Instance & Type & Layer 1 Param. & Reward Fn (Layer 1) & Layer 2 Param. & Reward Fn (Layer 2) & Size\\
\hline
1 & \ref{sec:comb_network} & SW-UCB & Mean Return ($w_f$=100) & CADTS & Sharpe Index ($w_f$=60) & k=4 \\
2 & \ref{sec:comb_network} & ADTS & Mean Return ($w_f$=100) & CADTS & Sharpe Index ($w_f$=60) & k=4 \\
3 & \ref{sec:2_layer_network} & ADTS & Mean Return ($w_f$=100) & ADTS & Sharpe Index ($w_f$=60) & k=4 \\
4 & \ref{sec:2_layer_network} & ADTS & Mean Return ($w_f$=100) & ADTS & Sharpe Index ($w_f$=60) & k=10 \\
5 & \ref{sec:2_layer_network} & ADTS & Mean Return ($w_f$=100) & ADTS & Sharpe Index ($w_f$=60) & k=15 \\
\hline
\label{tab:sp_portfolio_setup}
\end{tabular}
\end{table}

\subsection{Portfolio Selection Robustness Experiment}
\label{sec:s_p_portfolio_drift}
To verify the robustness of the Bandit Networks instances presented in Section~\ref{sec:s_p_portfolio} Table~\ref{tab:sp_portfolio_setup}, we incrementally remove a rank of the nine best stocks in cumulative returns. For the number of the top removed stocks, we define the variable $M$. Table~\ref{tab:sp_portfolio_robustness_setup} displays the experiment setup containing the simulation steps and the list of removed stocks at each step. We investigate financial metrics such as Cumulative Returns, Sharpe Index, and Maximum Drawdown for each step and Bandit Network instance and compare the results with the S\&P index and the CAPM portfolio model.

The results of these experiments yield insight into understanding the drift evolution of each studied Bandit Network instance and evaluate their dependencies to outlier performing stocks. In the next section, we present the results of three offered experiments in this study.

\begin{table}[ht]
\caption{S\&P Portfolio Optimization Robustness - Experiment Setup}
\centering
\begin{tabular}{cl}
\hline
Step & Stocks Removed\\
\hline
1 & -\\
2 & [NVDA]\\
3 & [NVDA, AVGO]\\
4 & [NVDA, AVGO, TSLA]\\
5 & [NVDA, AVGO, TSLA, LLY]\\
6 & [NVDA, AVGO, TSLA, LLY, GE]\\
7 & [NVDA, AVGO, TSLA, LLY, GE, AMD]\\
8 & [NVDA, AVGO, TSLA, LLY, GE, AMD, AAPL]\\
9 & [NVDA, AVGO, TSLA, LLY, GE, AMD, AAPL, XOM]\\
10 &[NVDA, AVGO, TSLA, LLY, GE, AMD, AAPL, XOM, GOOG]\\
\hline
\label{tab:sp_portfolio_robustness_setup}
\end{tabular}
\end{table}






\section{Results}\label{sec5}
The results of our experiments to assess the performance of the non-stationary bandits and the bandit network instances are presented in this section. As described in the previous section, we conducted three experiments: the stock picking experiment, the portfolio optimization experiment, and the portfolio robustness experiment.

\subsection{Stock Picking Experiment}
\label{sec:s_p_stock_results}
The first set of experiment results is towards the S\&P Stock Picking problem. The results are divided into three parts. In the first part, we analyze the learning characteristics of the ADTS against the bandit algorithms. Secondly, we obtain the financial metrics of each algorithm. Finally, we simulate the drift effect after applying shock in the top-performing stock of our S\&P set.

\subsubsection{Regret Analysis}

\begin{figure}[H]
\centering
\includegraphics[width=0.9\textwidth]{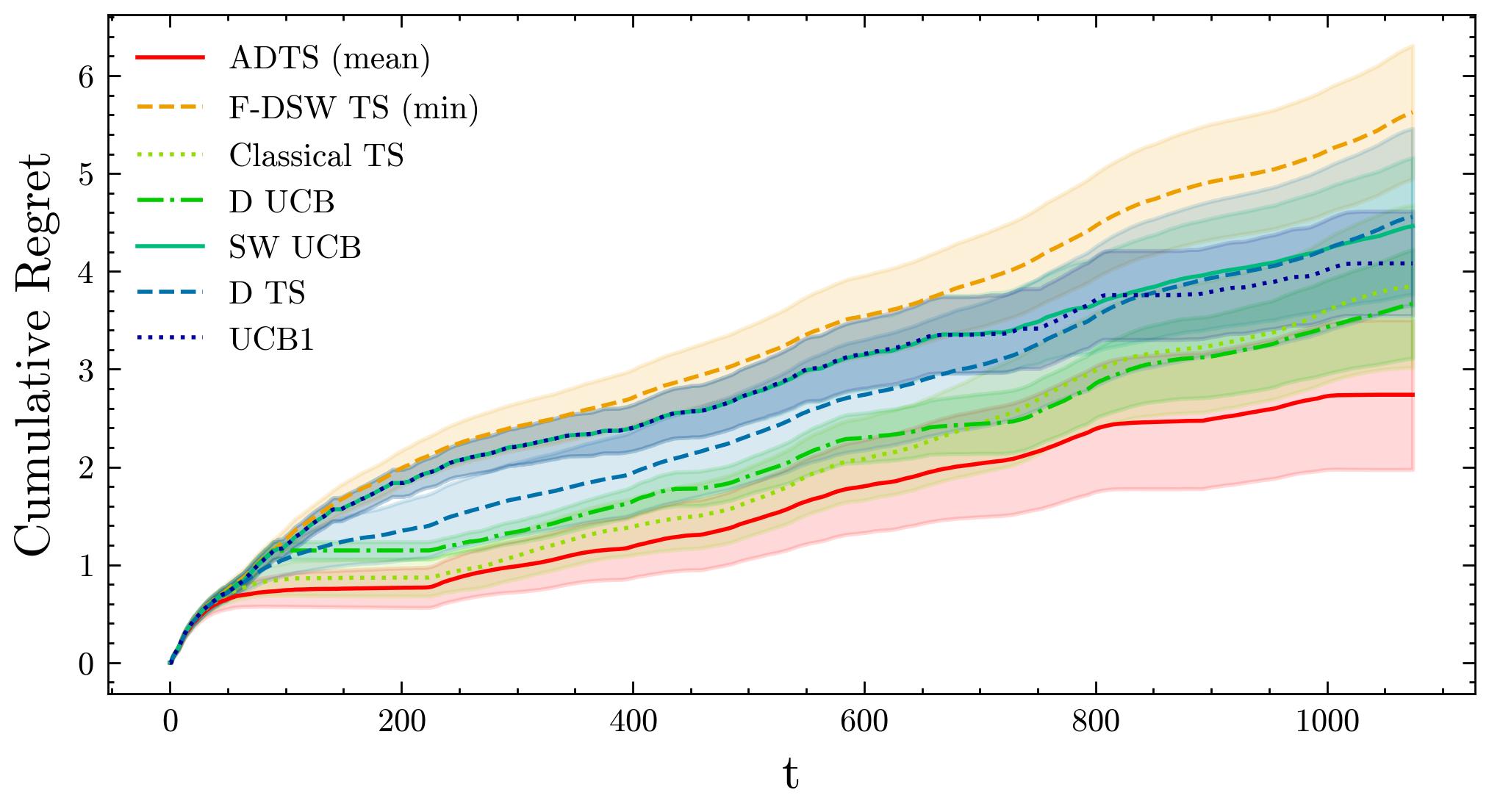}
\caption{Cumulative regret analysis, comparing the algorithms present in Table~\ref{tab:sp_stock_picking_setup} (based on 30 simulations for each policy)}
\label{fig:6.1_regrets}
\end{figure}

Figure \ref{fig:6.1_regrets} shows the cumulative regrets obtained according to the bandit algorithms present in Section~\ref{sec:s_p_stock} and Table~\ref{tab:6.1_regrets} summarizes the results to help the reader to understand the differences. 

The proposed ADTS is the one with the most prominent capability of detecting abrupt changes while presenting the lowest cumulative regret ($3.2 \pm 0.7$) (as per the red line with credible intervals). The top three rankings are completed with D UCB ($3.6 \pm 0.5$) and Classical TS ($3.9 \pm 0.7$). It draws attention that F-DSW TS (\citeauthor{cavenaghi2021nonstationary}, \citeyear{cavenaghi2021nonstationary}), the variant on which ADTS is based, presents the worst cumulative regret ($5.6 \pm 0.6$) when applied to the S\&P Stock Picking problem.

\begin{table}[h]
\caption{Comparison of Cumulative Regrets with 95\% Confidence Intervals (based on 30 simulations for each policy)}
\centering
\begin{tabular}{lcc}
\hline
Algorithm & Mean Cumulative Regret (95\% Conf.) \\
\hline
\textbf{ADTS (mean)} & \textbf{$2.7 \pm 0.8$} \\
D UCB & $3.7 \pm 0.6$ \\
Classical TS & $3.8 \pm 0.8$ \\
UCB1 & $4.1 \pm 0.5$ \\
SW UCB & $4.5 \pm 0.7$ \\
D TS & $4.6 \pm 0.9$ \\
F-DSW TS (min) & $5.6 \pm 0.7$ \\
\hline
\end{tabular}
\label{tab:6.1_regrets}
\end{table}

\subsubsection{Financial Metrics Analysis}

For a financial performance evaluation of the non-stationary Bandits policies, we investigate the following metrics: Return, Sharpe Ratio, Drawdown, Win Rate, and Sortino Ratio. Return quantifies profitability, while the Sharpe Ratio assesses risk-adjusted returns. Drawdown measures maximum loss, Win Rate indicates success frequency, and Sortino Ratio evaluates downside risk. These metrics collectively provide a comprehensive overview of a strategy's performance and risk profile. Results are stored in Table~\ref{tab:6.2_sp_picking_history}.

\begin{figure}[H]
\centering
\includegraphics[width=0.9\textwidth]{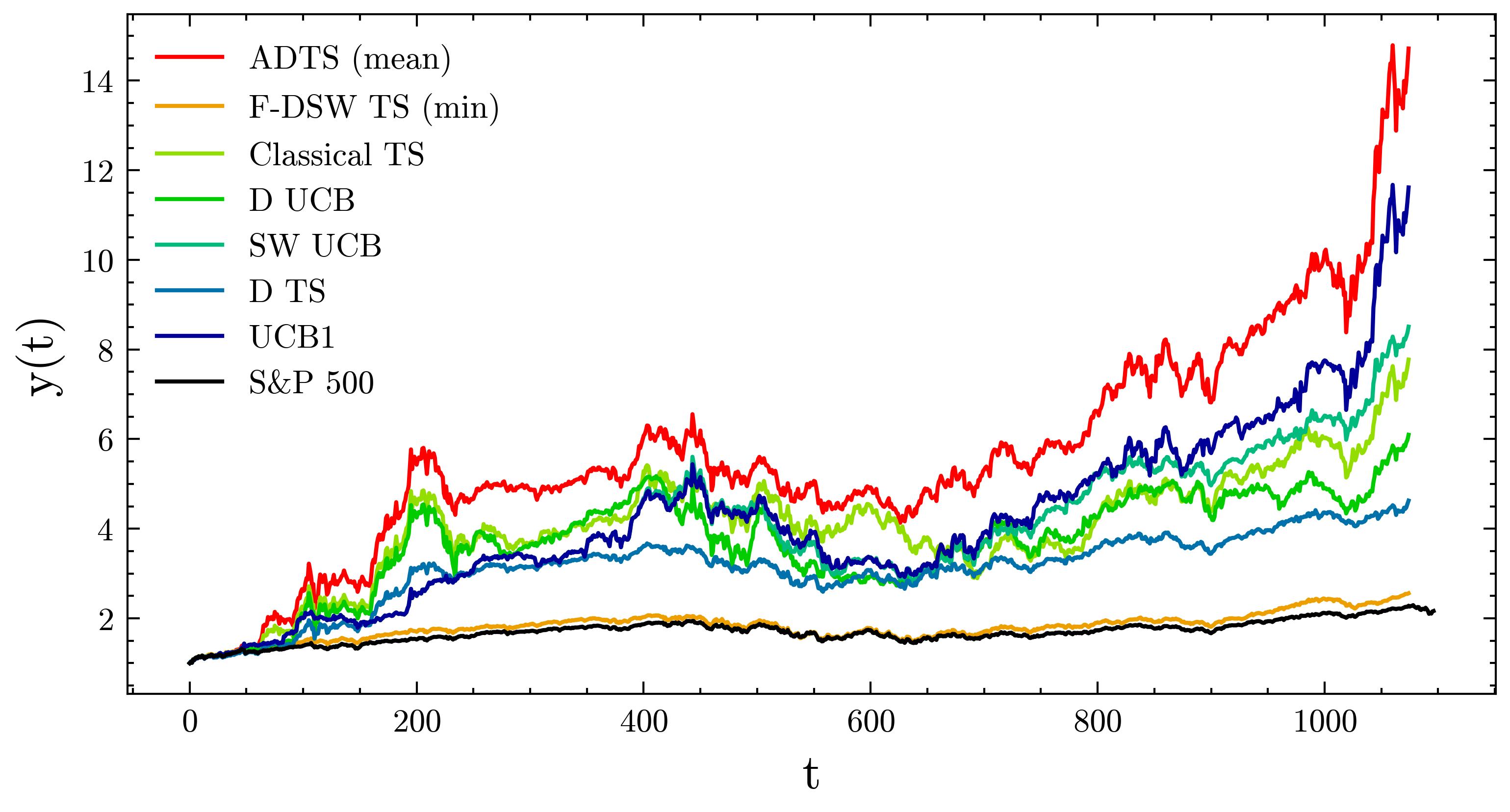}
\caption{Cumulative daily returns, comparing the algorithms present in Table~\ref{tab:sp_stock_picking_setup} (based on 30 simulations for each policy)}
\label{fig:6.2_cumulative_returns}
\end{figure}

Figure~\ref{fig:6.2_cumulative_returns} illustrates the cumulative returns obtained for each bandit algorithm in the experiment and the S\&P Index. Not only does the ADTS present the highest stock picking capability, but it transforms it into considerably better returns compared to the other Bandit policies or the S\&P 500 Index itself. When analyzing the Sharpe Ratio, UCB-1 is leading the metrics, followed by SW UCB and ADTS, in third. These three instances also stay at the top for the Sortino Ratio.

Compared to the S\&P 500 Index, in terms of returns, all the bandits policies present superior performance than the S\&P 500 Index. It is worth mentioning that all the policies can select one stock at a time, so maybe the higher drawdowns compared to the index, which is an aggregation of various stocks, are justified by this asymmetry.

\begin{table}[h]
\centering
\caption{Policies financial performance metrics}
\begin{tabular}{lcccccc}
\hline
Policy & Total Return & Sharpe Ratio & Drawdown & Win Rate & Sortino Ratio \\
\hline
\textbf{ADTS (mean)} & \textbf{13.72} & $1.76$ & $2.42$ & $0.55$ & $0.16$\\
UCB1 & $10.62$ & \textbf{1.91} & $2.52$ & $0.55$ & \textbf{0.19}\\
SW UCB & $7.51$ & $1.86$ & $2.78$ & $0.54$ & $0.18$\\
Classical TS & $6.77$ & $1.35$ & $2.52$ & $0.54$ & $0.12$\\
D UCB & $5.10$ & $1.21$ & $2.45$ & $0.55$ & $0.11$\\
D TS & $3.63$ & $1.51$ & $1.08$ & $0.54$ & $0.13$\\
F-DSW TS (min) & $1.56$ & $1.26$ & $0.59$ & $0.55$ & $0.11$\\
S\&P 500 & $1.20$ & $1.12$ & \textbf{0.49} & $0.54$ & $0.10$\\
\hline
\end{tabular}
\label{tab:6.2_sp_picking_history}
\end{table}

\subsubsection{Drift Analysis}

\begin{figure}[H]
\centering
\includegraphics[width=0.9\textwidth]{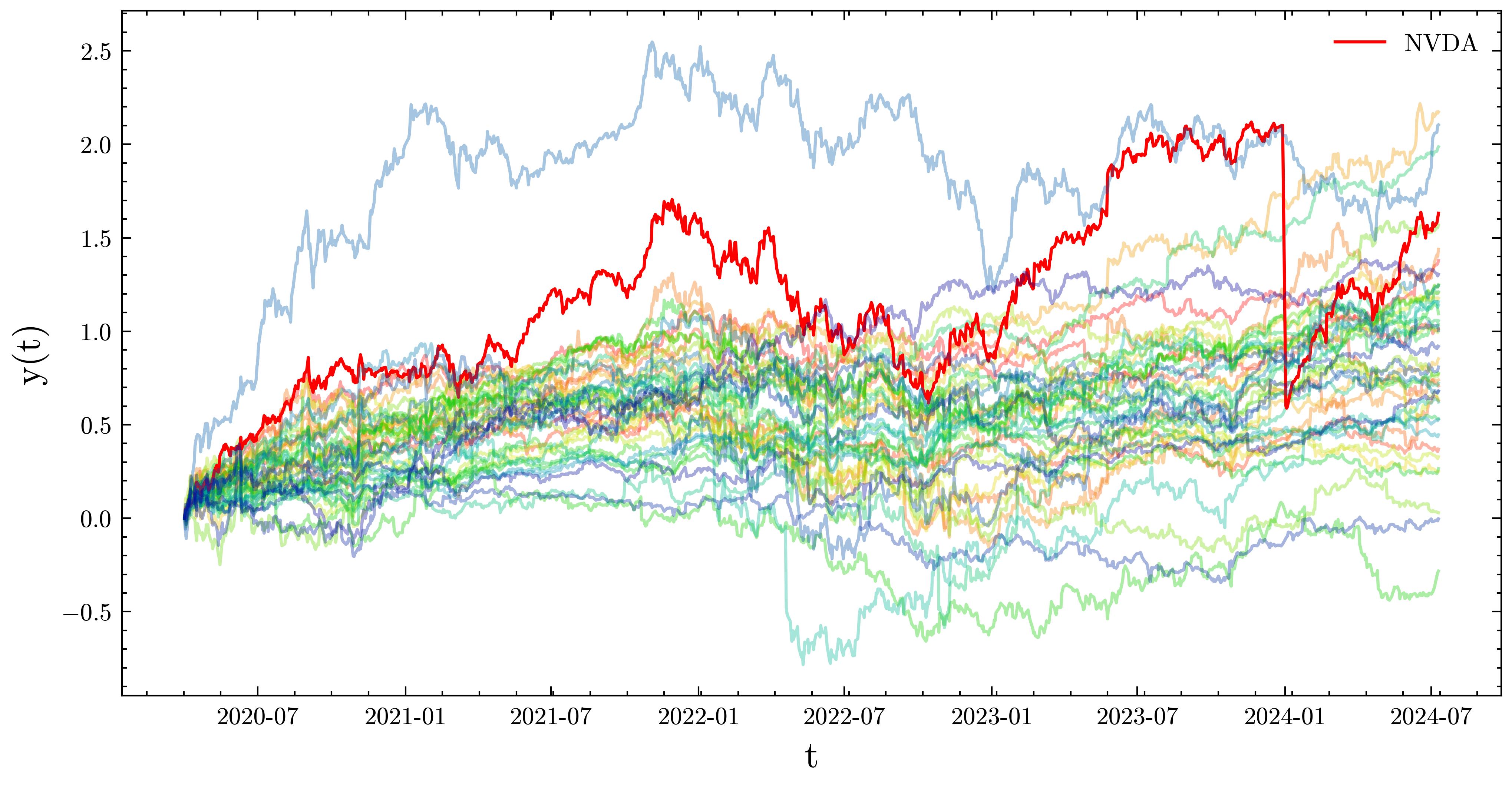}
\caption{Imposed drift to the best stock: NVDA}
\label{fig:6.3_nvda_drift}
\end{figure}

\begin{figure}[H]
    \centering
    \begin{subfigure}[b]{\textwidth}
        \centering
        \includegraphics[width=0.8\textwidth]{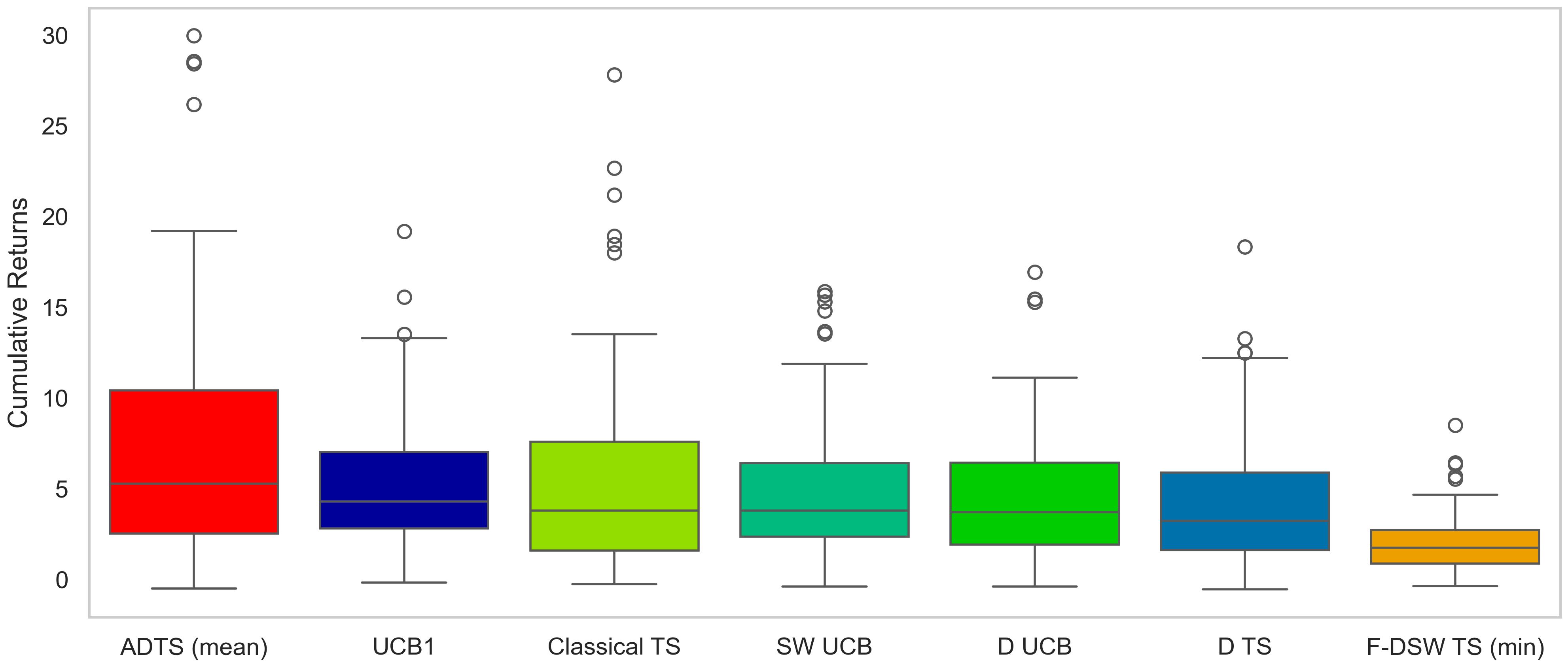}
        \caption{Cumulative returns}
    \end{subfigure}
    \begin{subfigure}[b]{\textwidth}
        \centering
        \includegraphics[width=0.8\textwidth]{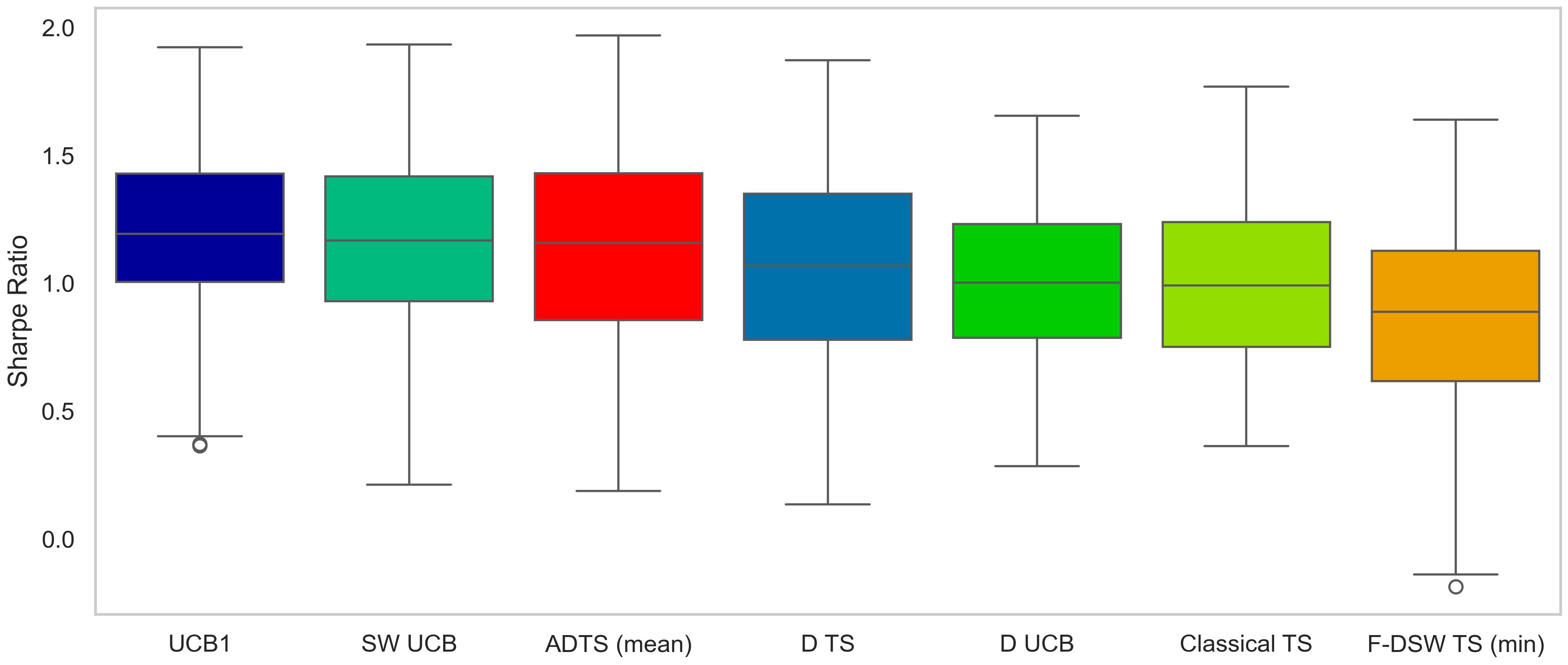}
        \caption{Sharpe Ratio}
    \end{subfigure}
    \begin{subfigure}[b]{\textwidth}
        \centering
        \includegraphics[width=0.8\textwidth]{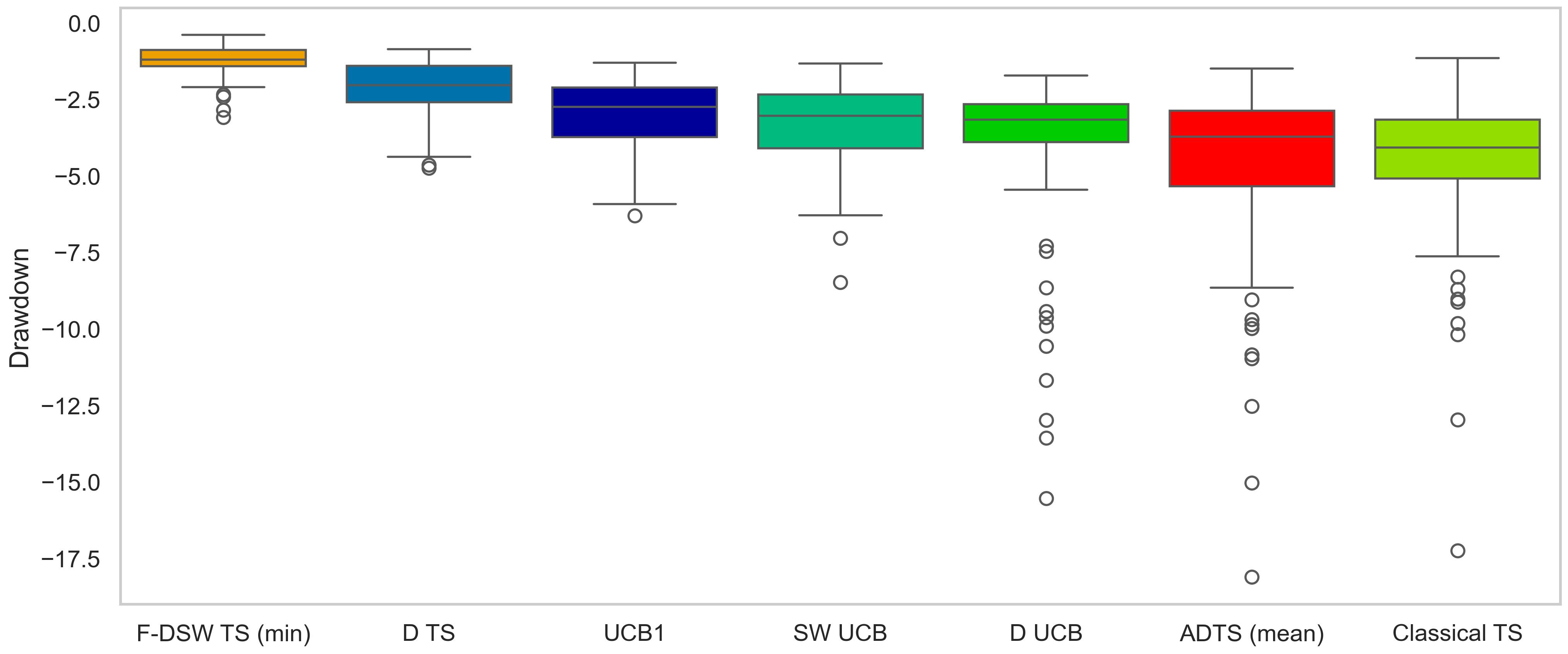}
        \caption{Drawdown}
    \end{subfigure}
    \caption{Drift analysis for the stock (30 simulations for each algorithm)}
    \label{fig:6.4_drift_analysis}
\end{figure}

For analyzing the concept drift in the selected set of S\&P stocks, we imposed an artificial shock to the best-performing ticker, the NVDA stock in January 2024. Figure~\ref{fig:6.3_nvda_drift} illustrates the change.
With this transformation, we repeated the S\&P Stock Picking problem for all the analyzed bandit algorithms, thirty (30) simulations for each policy. Results are shown in Figure~\ref{fig:6.4_drift_analysis}. After the applied drift, the ADTS is the policy that presents the highest cumulative median returns ($5.27$), followed by UCB-1 ($4.29$), which represents an increase of $22.8\%$. In terms of the Sharpe Ratio UCB-1 is leading ($1.20$), closely followed by ADTS ($1.17$). Being the worst performer policy in cumulative returns and Sharpe Ratio, the F-DSW TS is leading the rank for presenting the best risk behavior, given its Drawdown of $1.19$. On the other hand, our proposed policy, ADTS, had the second highest median Drawdown ($3.70$).

\subsection{Portfolio Optimization Experiment}
\label{sec:s_p_portfolioresults}
In the second experiment, we evaluate the performance of different Bandit Networks instances in the Portfolio Optimization Problem given a set of S\&P stocks. The results are split into two parts. In the first part, we analyze the learning characteristics of the Bandit Networks instances designed in Table~\ref{tab:sp_portfolio_setup}. Finally, we investigate the financial metrics of each network instance and compare them to classical portfolio allocation models and the S\&P Index.

\subsubsection{Regret Analysis}

\begin{figure}[H]
\centering
\includegraphics[width=0.9\textwidth]{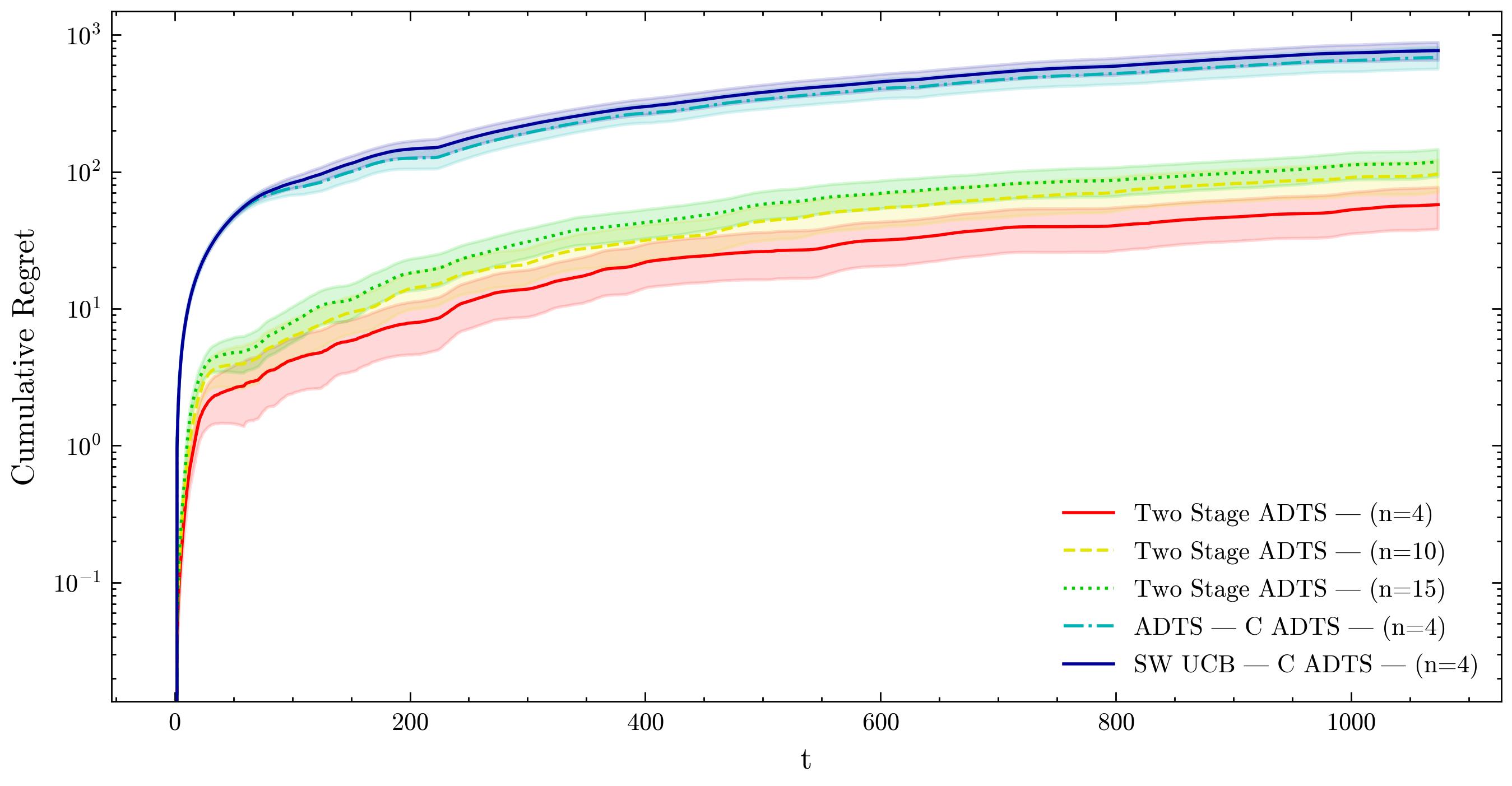}
\caption{Cumulative regret analysis, comparing the network instances present in Table~\ref{tab:sp_portfolio_setup}.}
\label{fig:6.2_regrets}
\end{figure}

Figure \ref{fig:6.2_regrets} shows the cumulative regrets in the log-scale for the y-axis obtained for the studied network instances. Table~\ref{tab:6.2_regrets} summarizes the results to help the reader to understand the differences. 

Comparing the regret results, the network instance 3 (Two-Layer ADTS, with $n=4$) stands out as the best learning configuration showing the lowest cumulative regret ($57.5 \pm 19.5$). The top three are completed by the other two instances derived from Section~\ref{sec:2_layer_network}, instances 4 and 5. There is a clear separation between the network instances proposed by Section~\ref{sec:comb_network}. Instance 1 presents the worst mean cumulative regret value ($767.2 \pm 113.6$), technically tied with instance 2 ($685.3 \pm 122.5$).

\begin{table}[ht]
\caption{Comparison of Cumulative Regrets with 95\% Confidence Intervals (based on 30 simulations for each policy)}
\centering
\begin{tabular}{lcc}
\hline
Bandit Network Instance & Mean Cumulative Regret (95\% Conf.) \\
\hline
Two Layer ADTS (n=4) & $57.5 \pm 19.5$ \\
Two Layer ADTS(n=10) & $96.4 \pm 25.3$ \\
Two Layer ADTS (n=15) & $118.8 \pm 26.9$ \\
ADTS $|$ CADTS (n=4) & $685.3 \pm 122.5$ \\
SW UCB $|$ CADTS (n=4) & $767.2 \pm 113.6$ \\
\hline
\end{tabular}
\label{tab:6.2_regrets}
\end{table}

\subsubsection{Financial Metrics Analysis}

\begin{figure}[H]
\centering
\includegraphics[width=0.9\textwidth]{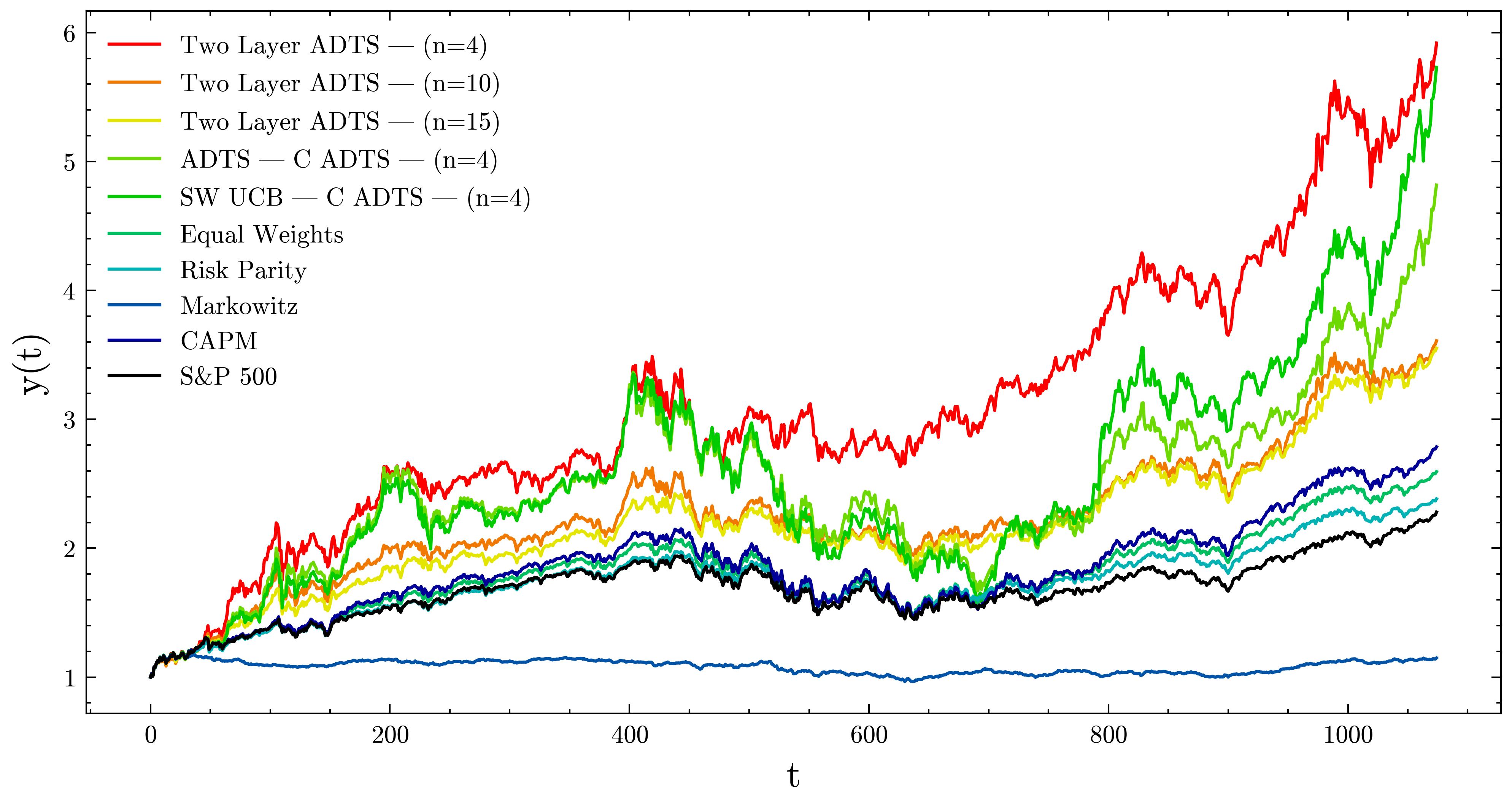}
\caption{Cumulative daily returns, comparing the network instances present in Table~\ref{tab:sp_portfolio_setup} (based on 30 simulations for each policy) and classical portfolio optimization models.}
\label{fig:6.3_sp_portfolio_returns}
\end{figure}

We move to analyze the financial metrics obtained for the bandit networks, comparing them with classical portfolio models. Results are stored in Table~\ref{tab:6.2_sp_portfolio_history}. Figure~\ref{fig:6.3_sp_portfolio_returns} illustrates the payoff chart of each of the bandit network instances results and the classical portfolio models and how they compare to the S\&P index.

The Two-Stage ADTS $(n=4)$ instance is the one with the most prominent results of cumulative returns ($4.92$), Sharpe and Sortino Ratios ($1.59$ and $0.14$, respectively).  

Sharpe Ratio value for Two-Stage ADTS $(n=15)$ slightly loses to the best instance. By selecting fifteen stocks simultaneously, the instance diversifies risks, as suggested by the smallest drawdown metric of the bandit networks instances ($0.55$).

Contrary to the cumulative regrets suggestions, SW UCB $|$ CADTS $(n=4)$ and ADTS $|$ CADTS $(n=4)$ stands in second and third when taking the cumulative returns, although compromising their Sharpe Ratio having higher risks than the other three remaining instances.

Compared to the classical portfolio models, nominally CAPM, Equal Weights, Risk parity, Markovitz as well as the S\&P 500 Index, in terms of returns, all the bandits network instances present superior performance. In this aspect, the cumulative returns of Two-Stage ADTS $(n=4)$ is 168\% higher than the CAPM, where the last is the best-performing classical model. The worst instance, Two-Stage ADTS $(n=15)$, presents cumulative returns 42\% higher than the best classical model, the CAPM, 2.55 against 1.79, respectively. 

The pattern persists when it comes to the Sharpe Ratio. The best network instance in this criteria, Two-Stage ADTS $(n=4)$, presents a Sharpe Ratio 20\% higher than the best classical model, the Equal Weights, 1.59 against 1.32, respectively. The other instances also present superior values when compared to Equal Weights, except for ADTS $|$ CADTS $(n=4)$ ($1.25$), which marginally loses to equal weights and risk parity models.

\begin{table}[ht]
\centering
\caption{Policies financial performance metrics}
\begin{tabular}{lccccc}
\hline
Network Instance & Return & Sharpe & Drawdown & Win Rate & Sortino \\
\hline
Two Layer ADTS (n=4) & \textbf{4.92} & \textbf{1.59} & $0.90$ & $0.55$ & \textbf{0.14}\\
SW UCB $|$ CADTS (n=4) & $4.73$ & $1.37$ & $1.78$ & $0.55$ & $0.13$\\
ADTS $|$ CADTS (n=4) & $3.82$ & $1.30$ & $1.70$ & $0.56$ & $0.12$\\
Two Layer ADTS (n=10) & $2.61$ & $1.47$ & $0.71$ & $0.56$ & $0.13$\\
Two Layer ADTS (n=15) & $2.55$ & $1.58$ & $0.55$ & $0.56$ & $0.14$\\
CAPM & $1.79$ & $1.29$ & $0.69$ & $0.55$ & $0.12$\\
Equal Weights & $1.59$ & $1.32$ & $0.59$ & $0.56$ & $0.12$\\
Risk Parity & $1.38$ & $1.31$ & $0.50$ & $0.54$ & $0.11$\\
S\&P 500 & $1.28$ & $1.17$ & $0.49$ & $0.54$ & $0.10$\\
Markowitz & $0.15$ & $0.40$ & \textbf{0.22} & $0.53$ & $0.04$\\
\hline
\end{tabular}
\label{tab:6.2_sp_portfolio_history}
\end{table}

\subsection{Portfolio Selection Robustness Experiment}
\label{sec:s_p_portfoliorobustness}
\begin{figure}[H]
\centering
\includegraphics[width=0.9\textwidth]{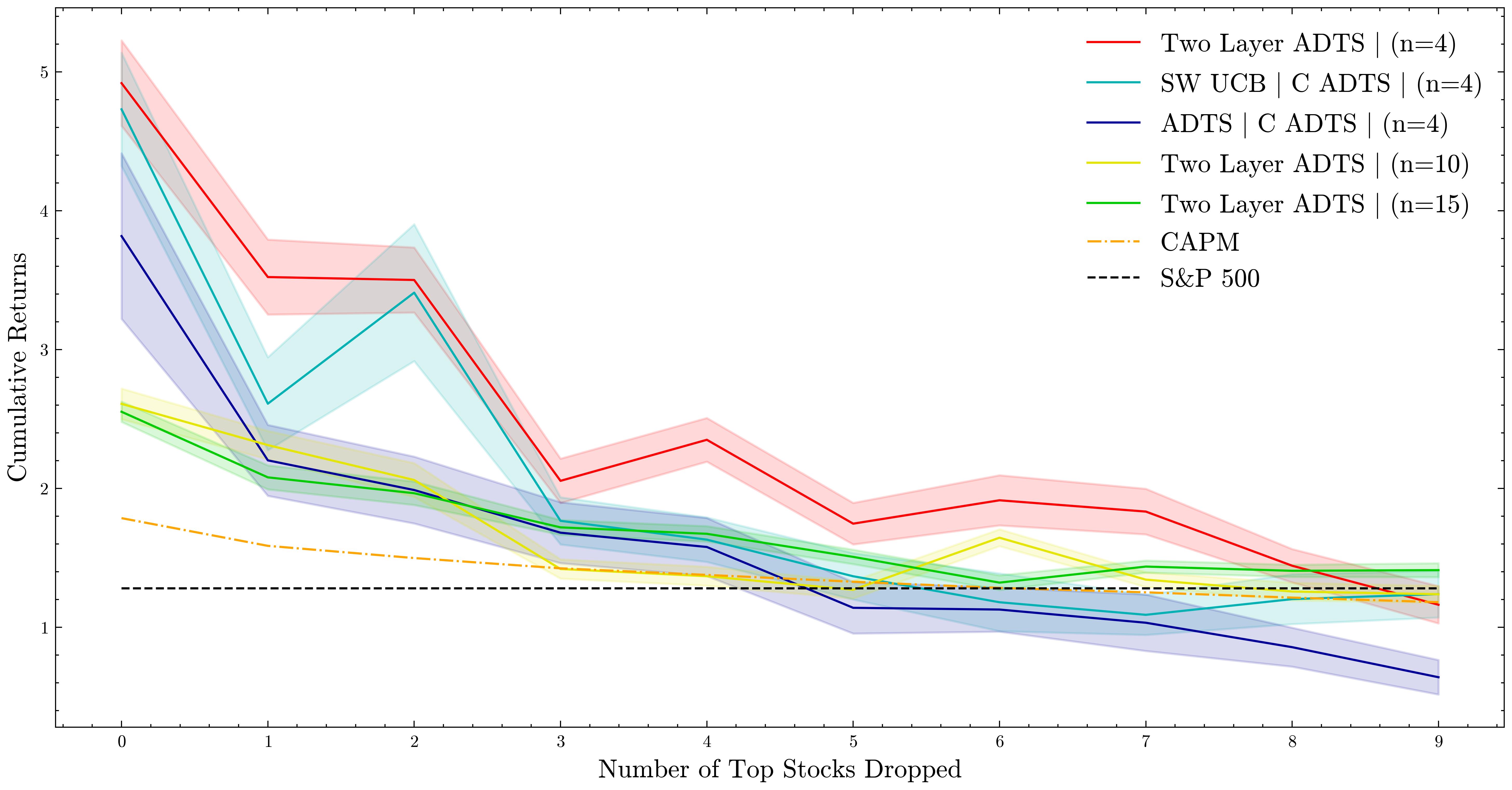}
\caption{Cumulative Returns drift analysis of bandit networks instances, after incrementally removing the best stocks in cumulative returns given in Table~\ref{tab:sp_portfolio_robustness_setup}.}
\label{fig:6.3_robustness_returns}
\end{figure}

\begin{figure}[H]
\centering
\includegraphics[width=0.9\textwidth]{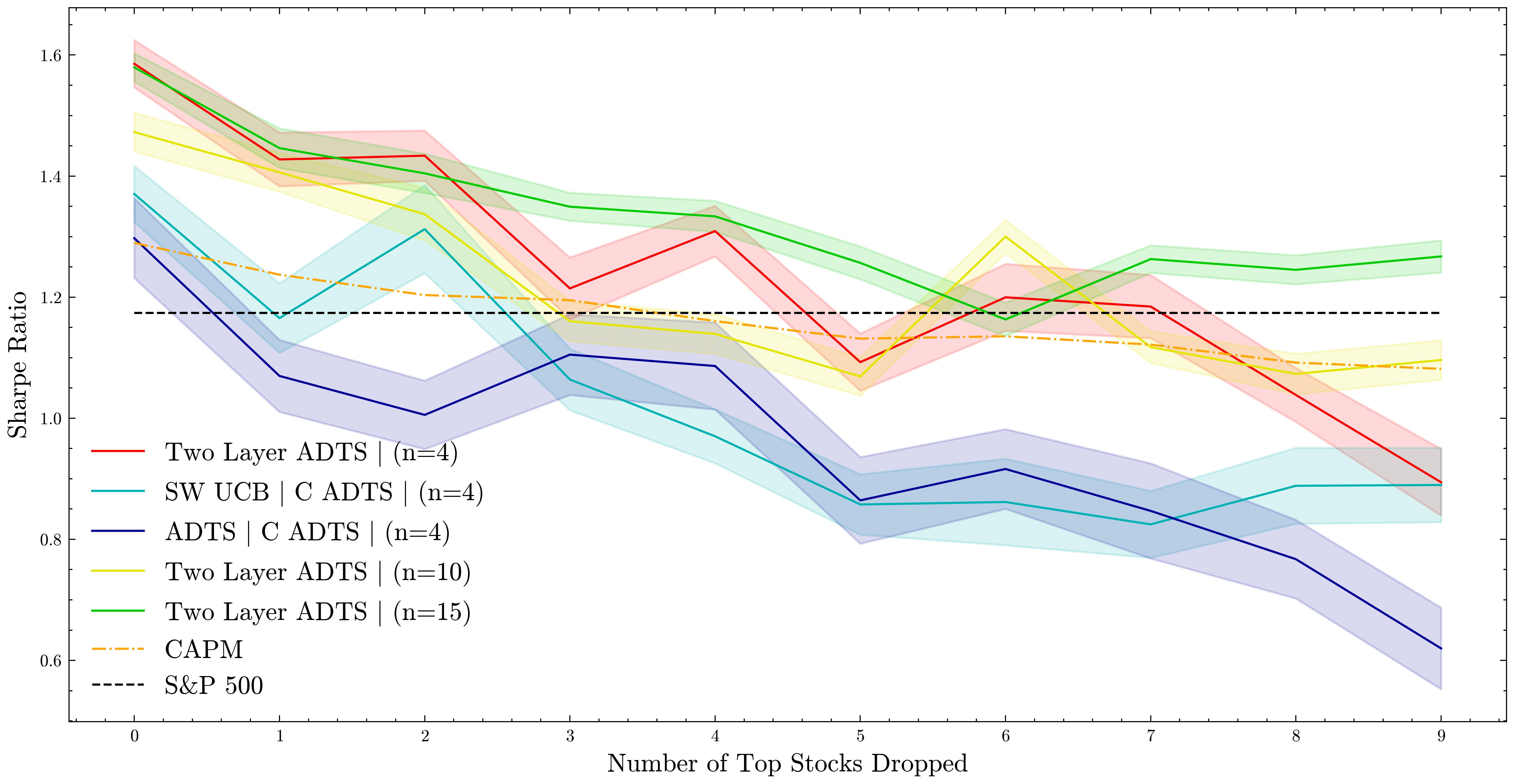}
\caption{Sharpe Ratio drift analysis of bandit networks instances, after incrementally removing the best stocks in cumulative returns given in Table~\ref{tab:sp_portfolio_robustness_setup}.}
\label{fig:6.3_robustness_sharpe}
\end{figure}

\begin{figure}[H]
\centering
\includegraphics[width=0.9\textwidth]{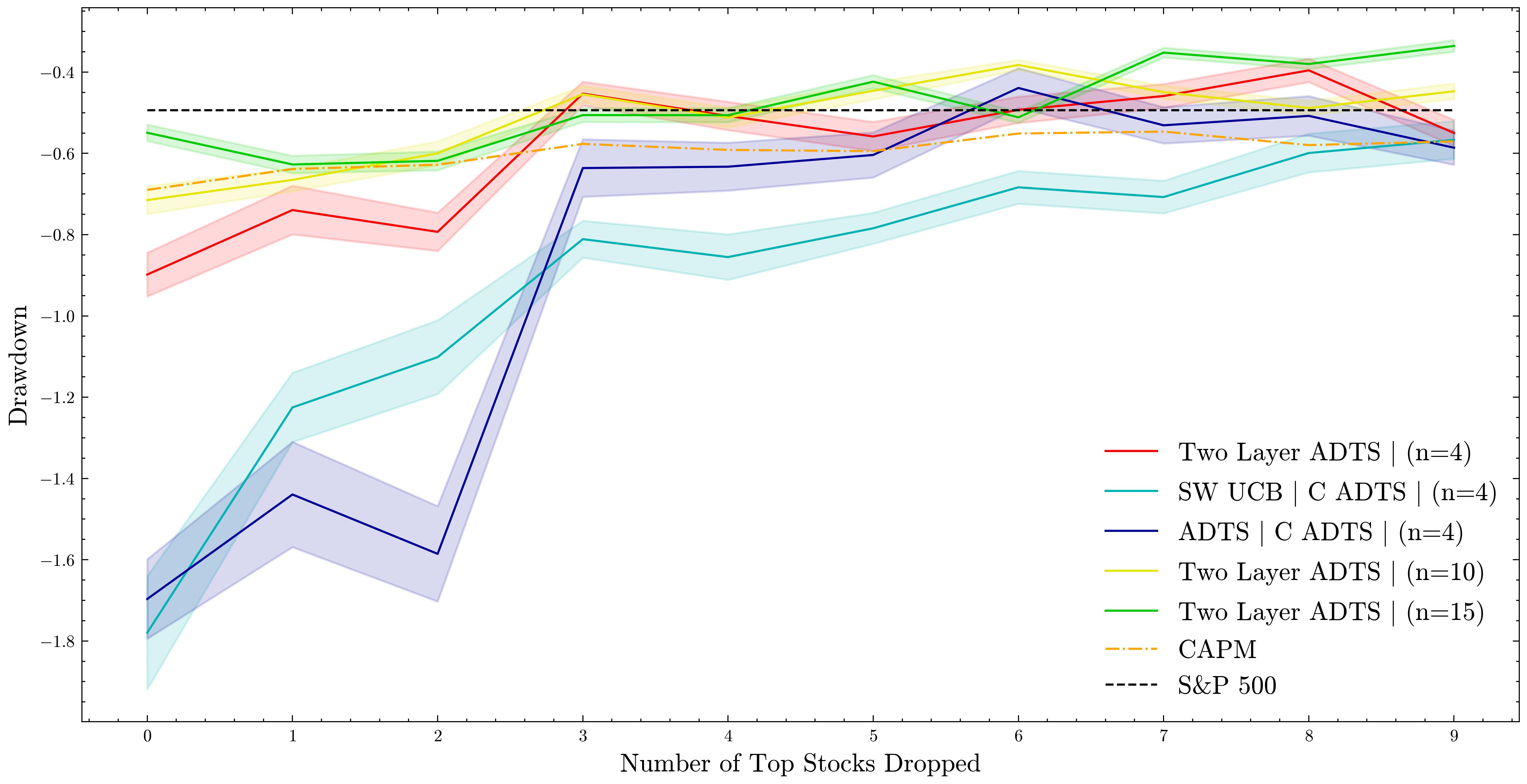}
\caption{Drawdown drift analysis of bandit networks instances, after incrementally removing the best stocks in cumulative returns given in Table~\ref{tab:sp_portfolio_robustness_setup}.}
\label{fig:6.3_robustness_drawdown}
\end{figure}

To finish the set of experiments, we present the results depicted in the experiment setup presented in Section~\ref{sec:s_p_portfolio_drift}. To quantify how much each of the bandit network instances sustains their performance after removing the best stocks, we instantiate three types of financial metrics: i) Cumulative Returns (return), ii) Sharpe Ratio (return adjusted to risk), and iii) Drawdown (risk). Figures~\ref{fig:6.3_robustness_returns}-\ref{fig:6.3_robustness_drawdown} show line plots of the three mentioned financial metrics as a function of the number of the best stocks ($M$) shown in the experiment setup section. Table~\ref{tab:6.3_robustness_metrics} consolidates the three analyzed metrics for each methodology, storing the values when $M=0, 4, 9$. We compare the bandit network instances with the CAPM model and the S\&P Index.

As shown in the table, the three instances with $n=4$ present higher drifts, which is logical due to the less diversification when compared to the instances with $n=10$ and $n=15$. 

For Cumulative Returns, the network that uses two layers of ADTS (red line) maintains the highest values until the number of the dropped best reaches eight. On the other hand, the networks that use CADTS in the last layer (Section~\ref{sec:comb_network}) do not manage the same capability, as they start to lose to the $n=10$ and $n=15$ instances, and even the CAPM and S\&P Index after removing six best stocks. The Two Layer ADTS ($n=15$), green line in the charts, demonstrates the highest bandit network capability of maintaining the cumulative returns after $M=9$, preserving a value of 1.41, 19\% higher than CAPM and 10\% higher than the S\&P Index. Overall, the analyzed bandit networks demonstrate higher cumulative return drift when compared to CAPM and S\&P Index.

For the Sharpe Ratio, similar trends can be observed. The two-layer ADTS ($n=15$) showcases the lowest drift values amongst all the bandit network instances, demonstrating comparable stability to the CAPM model. In fact, after $M=9$ it presents the highest value (1.27), 17\% higher than the CAPM model, the second best. Similar to cumulative returns, in this criteria the higher drift values are also observed for the bandit network instances.

Finally, for the Drawdown metric, the two-layer ADTS ($n=15$) presents the top three less risky choices, including the CAPM and S\&P Index. After the number of best stocks removed is equal or superior to seven, the instance presents the lowest value. This less risky behavior is certainly helping it to sustain the highest Sharpe Ratio, given the fact that its Cumulative Returns values are modest though stable, compared to the other instances.

\begin{table}[ht]
\centering
\caption{Robustness Analysis}
\begin{tabular}{lccccc}
\hline
Network Instance & M=0 & M=4 & M=9 & Total Drift (M=0 to M=9)\\
\hline
\textbf{Cumulative Return} & & & &\\
Two Layer ADTS (n=4) & 4.92 & 1.75 & 1.16 & 76.4\%  \\
SW UCB $|$ CADTS (n=4) & 4.73 & 1.37 & 1.24 & 73.8\%  \\
ADTS $|$ CADTS (n=4) & 3.82 & 1.14 & 0.64 & 83.2\%  \\
Two Layer ADTS (n=10) & 2.61 & 1.27 & 1.24 & 52.5\%  \\
Two Layer ADTS (n=15) & 2.55 & 1.51 & 1.41 & 44.7\%  \\
CAPM & 1.79 & 1.33 & 1.18 & 33.9\%  \\
S\&P 500 & 1.28 & 1.28 & 1.28 & 0.0\%  \\
 & & & &\\
\textbf{Sharpe Ratio} & & & &\\
Two Layer ADTS (n=4) & 1.59 & 1.09 & 0.89 & 43.6\%  \\
SW UCB $|$ CADTS (n=4) & 1.37 & 0.86 & 0.89 & 35.1\%  \\
ADTS $|$ CADTS (n=4) & 1.3 & 0.86 & 0.62 & 52.2\%  \\
Two Layer ADTS (n=10) & 1.47 & 1.07 & 1.1 & 25.6\%  \\
Two Layer ADTS (n=15) & 1.58 & 1.26 & 1.27 & 19.8\%  \\
CAPM & 1.29 & 1.13 & 1.08 & 16.1\%  \\
S\&P 500 & 1.17 & 1.17 & 1.17 & 0.0\%  \\
\hline
\end{tabular}
\label{tab:6.3_robustness_metrics}
\end{table}

\section{Discussion}\label{sec6}
This section discusses the outcomes of our experiments on stock picking, portfolio optimization, and portfolio robustness. These experiments evaluate the performance of the newly introduced ADTS algorithm, as well as the novel concept of bandit networks presented in this work. With this goal, we used a set of 44 stocks' historical daily returns of the S\&P index, starting from April 2020 to July 2024.

The stock-picking experiments reveal the superior performance of the proposed ADTS algorithm. Regret analysis demonstrates that ADTS achieves the lowest cumulative regret, significantly outperforming other bandit algorithms, including its predecessor F-DSW TS. Financial metrics further support its efficacy, with ADTS yielding the highest returns among the tested algorithms, and also showing commendable results in terms of Sharpe and Sortino ratios. Notably, all bandit policies surpass the S\&P 500 Index in returns, though they exhibit higher drawdowns, likely due to their single-stock selection constraint. Drift analysis, conducted by imposing a shock on the top-performing NVDA stock, underscores ADTS's robustness, maintaining high cumulative returns and a competitive Sharpe Ratio even under perturbations.

In portfolio optimization, the two-layer ADTS network with $n=4$ stocks stands out with the lowest cumulative regret and highest cumulative returns, Sharpe, and Sortino ratios. Compared to classical portfolio models like CAPM, Equal Weights, Risk Parity, Markowitz, and the S\&P 500 Index, all bandit network instances exhibit superior performance. Notably, the cumulative returns of the two-layer ADTS ($n=4$) are 168\% higher than CAPM, the best-performing classical model. Even the worst instance, two-layer ADTS ($n=15$), shows cumulative returns 42\% higher than CAPM.

This trend continues with the Sharpe Ratio, where the two-layer ADTS ($n=4$) is 20\% higher than Equal Weights, the best classical model in this regard. Other instances also surpass Equal Weights, except for ADTS $\mid$ CADTS ($n=4$), which slightly trails behind Equal Weights and Risk Parity. While other network instances, particularly those combining ADTS and SW UCB, show higher cumulative regrets, they still outperform classical models in returns. The two-layer ADTS networks with higher $n$ values (10 and 15) better diversify risks, as indicated by their lower drawdowns.

The robustness experiments corroborate the resilience of the two-layer ADTS networks, especially with higher $n$ values. For cumulative returns, the two-layer ADTS ($n=4$) maintains the highest values until eight top-performing stocks are removed. In contrast, networks using CADTS in the last layer start losing to the $n=10$ and $n=15$ instances, and even the CAPM and S\&P Index after removing six top stocks. The two-layer ADTS ($n=15$) demonstrates the highest robustness, maintaining cumulative returns 19\% higher than CAPM and 10\% higher than the S\&P Index after removing the nine best-performing stocks.

For the Sharpe Ratio, the two-layer ADTS ($n=15$) shows the lowest drift values among all bandit network instances and is 17\% higher than the CAPM model after $M=9$. For the Drawdown metric, the two-layer ADTS ($n=15$) is among the top three least risky options, along with CAPM and the S\&P Index, and presents the lowest drawdown when seven or more top stocks are removed. This low-risk behavior contributes to sustaining the highest Sharpe Ratio, highlighting the practical utility of the two-layer ADTS networks in maintaining portfolio performance amidst market fluctuations.


\section{Conclusion}\label{sec7}
This work introduced and evaluated the ADTS algorithm and the concept of bandit networks through a series of experiments on stock picking, portfolio optimization, and portfolio robustness, using historical daily returns of 44 S\&P 500 stocks from April 2020 to July 2024. The ADTS algorithm demonstrated superior performance, consistently achieving the lowest cumulative regret and highest returns, showcasing its effectiveness in both static and dynamic market conditions. The two-layer ADTS networks, particularly with $n=4$ and $n=15$, exhibited remarkable robustness and risk-adjusted returns, outperforming classical models such as CAPM and Equal Weights.

The stock-picking experiments highlighted the ADTS's ability to maintain high returns and competitive Sharpe Ratios even under concept drift. In the portfolio optimization results, the two-layer ADTS networks efficiently learned and adapted, yielding superior cumulative returns and risk metrics. The robustness analysis further validated the stability of these networks, especially with higher $n$ values, in maintaining performance amidst market fluctuations.

Future work could explore the application of ADTS and bandit networks to a broader range of financial instruments and market conditions. Additionally, enhancing the models to mitigate higher drawdowns observed in stock picking could further improve their practicality. Investigating the integration of alternative financial metrics and incorporating real-time adaptive mechanisms may also provide valuable insights for developing more resilient and adaptive financial decision-making tools.

\backmatter





\bmhead{Acknowledgements}
This work was partially supported by grant 2022/01524-2, São Paulo Research Foundation (FAPESP).

\section*{Declarations}


\begin{itemize}
\item \textbf{Funding.} This study was supported by the Fundação de Amparo a Pesquisa do Estado de São Paulo (FAPESP - Grant 2022/01524-2).
\item \textbf{Conflict of interest/Competing interests.} The authors have no relevant financial or non-financial interests to disclose.
\item \textbf{Author's contribution.} All the authors contributed to the study’s conception and design. Material preparation, data collection and analysis were performed by Gustavo Fonseca. The first draft of the manuscript was written by Gustavo Fonseca. Lucas Coelho and Paulo André Lima de Castro commented on previous versions of the manuscript. All the authors read and approved the final manuscript.
\end{itemize}

\noindent

\end{document}